\documentclass[twocolumn]{article}

\usepackage[a4paper, textwidth=7.2in, top=1in, bottom=1in]{geometry}

\usepackage[utf8]{inputenc}
\usepackage[round]{natbib}
\usepackage[colorlinks=true, allcolors=blue, hypertexnames=false]{hyperref}
\usepackage[font=small, labelfont=bf]{caption}

\usepackage[sc,osf]{mathpazo}    
\usepackage{graphicx}    
\usepackage{amsmath,amssymb}     
\usepackage{amsfonts}    
\usepackage{booktabs}    
\usepackage{multirow}    
\usepackage{microtype}   
\usepackage[misc]{ifsym} 

\usepackage{pstricks}
\usepackage{dsfont}
\usepackage{xy} 
\xyoption{all} 
\newcommand{\R}{\mathbb{R}}

\usepackage[capitalize]{cleveref}

\usepackage{titling}
\pretitle{\begin{flushleft}\rule{\textwidth}{0.8pt}\end{flushleft} \begin{center}\Huge\sc}
\posttitle{\par\end{center} 
\rule{\textwidth}{0.8pt} \vskip 0.5em}

\usepackage{authblk}

\usepackage{orcidlink}

\title{Low-dimensional embeddings\\of high-dimensional data}

\author[1,2,*]{Cyril de Bodt\,\orcidlink{0000-0003-2347-1756}\,}
\author[3,*]{Alex Diaz-Papkovich\,\orcidlink{0000-0002-2867-5494}\,}
\author[4]{Michael Bleher\,\orcidlink{0000-0002-7796-1665}\,} 
\author[5]{Kerstin Bunte\,\orcidlink{0000-0002-2930-6172}\,} 
\author[6,7,8]{Corinna~Coupette\,\orcidlink{0000-0001-9151-2092}\,} 
\author[9]{Sebastian Damrich\,\orcidlink{0000-0003-1394-6236}\,} 
\author[10,11]{Enrique Fita Sanmartin\,\orcidlink{0009-0004-8332-1164}\,}
\author[12]{Fred~A.~Hamprecht\,\orcidlink{0000-0003-4148-5043}\,} 
\author[13]{Em\H oke-\'Agnes Horv\'at\,\orcidlink{0000-0001-7709-1172}\,}
\author[14]{Dhruv Kohli\,\orcidlink{0000-0002-0306-913X}\,} 
\author[15]{Smita Krishnaswamy\,\orcidlink{0000-0001-5823-1985}\,}  
\author[2]{John~A.~Lee\,\orcidlink{0000-0001-5218-759X}\,} 
\author[16]{Boudewijn~P.~F.~Lelieveldt\,\orcidlink{0000-0001-8269-7603}\,} 
\author[17]{Leland McInnes\,\orcidlink{0000-0003-2143-6834}\,} 
\author[18]{Ian~T.~Nabney\,\orcidlink{0000-0001-7382-2855}\,} 
\author[19]{Maximilian~Noichl\,\orcidlink{0000-0003-4518-0837}\,} 
\author[20]{Pavlin~G.~Poli\v{c}ar\,\orcidlink{0000-0002-6462-9372}\,} 
\author[21]{Bastian Rieck\,\orcidlink{0000-0003-4335-0302}\,} 
\author[10,11]{Guy Wolf\,\orcidlink{0000-0002-6740-059X}\,} 
\author[14,+]{Gal~Mishne\,\orcidlink{0000-0002-5287-3626}\,}
\author[9,+]{Dmitry~Kobak\,\orcidlink{0000-0002-5639-7209}\,}

\affil[*]{Equal contribution}
\affil[+]{Equal contribution}
\affil[ ]{ }
\affil[1]{Department of Mathematics and Namur Research Institute for Complex Systems (naXys), University of Namur, Belgium}
\affil[2]{ICTEAM Institute, UCLouvain, Louvain-la-Neuve, Belgium}
\affil[3]{Data Science Institute, Brown University, Providence, Rhode Island, United States}
\affil[4]{Institute for Mathematics, Heidelberg University, Germany}
\affil[5]{University of Groningen, The Netherlands}
\affil[6]{Aalto University, Finland}
\affil[7]{Max Planck Institute for Informatics, Saarbrücken, Germany}
\affil[8]{Max Planck Institute for Tax Law and Public Finance, Munich, Germany}
\affil[9]{Hertie Institute for AI in Brain Health, University of T\"ubingen, Germany}
\affil[10]{Universit\'{e} de Montr\'{e}al, Canada}
\affil[11]{Mila, Montr\'{e}al, Canada}
\affil[12]{IWR, Heidelberg University, Germany}
\affil[13]{Northwestern University, Evanston, Illinois, United States}
\affil[14]{UC San Diego, California, United States}
\affil[15]{Yale University, New Haven, Connecticut, United States}
\affil[16]{Department of Radiology, Leiden University Medical Center, The Netherlands}
\affil[17]{Tutte Institute for Mathematics and Computing, Ottawa, Canada}
\affil[18]{University of Bristol, United Kingdom}
\affil[19]{Department of Philosophy and Religious Studies, Utrecht University, The Netherlands}
\affil[20]{Faculty of Computer and Information Science, University of Ljubljana, Slovenia}
\affil[21]{University of Fribourg, Switzerland}
\affil[ ]{ }
\affil[ ]{\Letter\: \normalfont{\texttt{dmitry.kobak@uni-tuebingen.de}}}

\usepackage{fancyhdr}
\begin{document}

\maketitle

\clearpage
\begin{abstract}
\noindent
Large collections of high-dimensional data have become nearly ubiquitous across many academic fields and application domains, ranging from biology to the humanities.
Since working directly with high-dimensional data poses challenges, the demand for algorithms that create low-dimensional representations, or \emph{embeddings}, for data visualization, exploration, and analysis is now greater than ever. 
In recent years, numerous embedding algorithms have been developed, and their usage has become widespread in research and industry. 
This surge of interest has resulted in a large and fragmented research field that faces technical challenges alongside fundamental debates, and it has left practitioners without clear guidance on how to effectively employ existing methods. 
Aiming to increase coherence and facilitate future work, 
in this review we provide a detailed and critical overview of recent developments, 
derive a list of best practices for creating and using low-dimensional embeddings, evaluate popular approaches on a variety of datasets, and discuss the remaining challenges and open problems in the field.
\end{abstract}



\section{Introduction}
\label{sec:intro}

At the heart of scientific inquiry lie curiosity and intuition. 
Scientists observe the world, detect patterns, create hypotheses, and make inferences.
The ever-increasing amount of data collected as part of virtually all scientific endeavors requires researchers to observe the data, detect patterns in the data, and formulate hypotheses about the data. 
Since John Tukey famously summarized these aims as \emph{exploratory data analysis} (EDA) in the 1970s \citep{tukey1977exploratory}, both the number of samples and the number of features in a typical dataset have increased by several orders of magnitude, motivating the development of new analytical approaches. 
One of these approaches is \emph{dimensionality reduction}.

Large collections of high-dimensional data have become widespread in almost every discipline, with samples of interest ranging from molecules in chemistry to publications in the digital humanities and single cells or organisms in biology. These samples are typically characterized by thousands of features, such as atomic properties, word frequencies, or gene-expression measurements, resulting in high-dimensional data.
However, empirical observations in many domains suggest that while the data are observed in high dimensions, the number of intrinsic dimensions that characterize the variability in the data is often much smaller. 
This has motivated the development of multiple dimensionality-reduction methods that learn low-dimensional representations, commonly referred to as \emph{embeddings}, of high-dimensional data. 

Low-dimensional embeddings typically have to sacrifice some high-dimensional information and distort the original data \citep{Welch1974LowerBO, Johnson1984ExtensionsOL, larsen2017JLoptimality, nonato2018multidimensional}.
This has motivated the development of many algorithms that make different trade-offs and preserve different properties of the data, resulting in a fast-growing and dynamic area of machine-learning research \citep{lee2007nonlinear,espadoto2019toward, nguyen2019ten,
armstrong2022applications, wang2023comparative, meilua2024manifold,wayland2024multiverse}.
The resulting methods are often used as a preprocessing step to reduce high-dimensional noise or to render downstream tasks such as clustering or prediction more computationally tractable. 

Embeddings have also become a powerful and popular tool in exploratory data analysis, enabling interpretable representations in two or three dimensions that are amenable to direct visualization and human interaction.
This has motivated the widespread adoption of low-dimensional embeddings across disciplines as different as 
single-cell biology \citep{kobak2019art, becht2019dimensionality}, 
human genetics \citep{diaz2019umap}, biochemistry \citep{durairaj2023uncovering}, 
philosophy \citep{noichl_modeling_2021}, 
metascience \citep{gonzalez-marquez2024},
and sports analytics \citep{garcia-aliaga_-game_2021}.

As the field evolves, it becomes increasingly important to not only develop better embedding methods but also establish best practices for the creation, evaluation, application, presentation, and documentation of low-dimensional embeddings in the scientific process.
In this review, following a reflection on the role of low-dimensional embeddings in the scientific enterprise (\cref{sec:epistemic}), we provide an overview of common approaches to dimensionality reduction and highlight popular algorithms representing each approach, focusing on their underlying motivations, advantages, and limitations (\cref{sec:algorithms}). We showcase selected algorithms on several example datasets from different research areas to illustrate the trade-offs involved (\cref{sec:showcasing}), offer specific guidance and best practices relevant to real-world applications (\cref{sec:best-practices}), and outline important challenges that remain in the field (\cref{sec:challenges}). 
The paper is co-authored by the participants of the Dagstuhl seminar 24122 on \emph{Low-Dimensional Embeddings of High-Dimensional Data}, held in March 2024, and builds on our seminar discussions~\citep{kobak_et_al:DagRep.14.3.92}.

\section{Epistemic roles of embeddings}
\label{sec:epistemic}

In scientific fields that collect large quantities of high-dimensional data, such as single-cell transcriptomics, embedding and visualizing data in 2D 
has become a nearly ubiquitous practice. This has generated some debate and controversy, with critics emphasizing inevitable distortions \citep{wang2023cannot,chariSpeciousArtSinglecell2023} and advocates emphasizing practical usefulness 
\citep{lause2024art}. 
Studies have also noted artifacts that can arise from visualizing data in low dimensions~\citep{diaconis2008horseshoes, morton2017uncovering, lebedev2019analysis}, and drew attention to frequent misuse of such embeddings \citep{jeon2025stop}. This raises an epistemological question: What role do low-dimensional embeddings play in the scientific endeavor? 

One answer is that (low-dimensional) embeddings fall into the category of \emph{exploratory} data analysis. EDA~\citep{tukey1977exploratory} was developed as a hypothesis-free approach to data analysis, complementing \textit{confirmatory} approaches such as hypothesis testing. Visualization techniques play a critical role in EDA because, as Tukey argued, ``the picture-examining eye is the best finder we have of the wholly unanticipated''~\citep{tukey_we_1980}. Hence, a low-dimensional visualization can reveal properties of the data that the researchers were not even considering \citep{yanai2020hypothesis}, sometimes referred to as \textit{unknown unknowns}.

Two-dimensional embeddings depicted as a scatter plot can be easily interpreted due to the \emph{Gestalt principles} of human perception~\citep{Koffka1935}, which characterize how humans group elements and recognize patterns. For example, the principle of proximity and the principle of common region state that objects placed near each other are perceived as similar, whereas those in separate areas are seen as distinct. 
Under this interpretation, rather than serving as tools for testing predefined hypotheses, embeddings serve a purpose similar to that of traditional box plots, scatter plots, and histograms, allowing researchers to inspect the data distribution. This inspection process can yield new insights and hypotheses that should subsequently be tested in a confirmatory framework\,---\,i.e., embeddings help with \emph{hypothesis generation}.

In EDA, the epistemic role of low-dimensional embeddings resembles that of \textit{microscopes} in biomedicine or \textit{telescopes} in astronomy: They can immediately reveal information that is otherwise hidden from the naked eye. 
In our own work, we have found 2D embeddings useful for guiding further research by exposing unexpected phenomena, such as clusters of retracted papers in collections of abstracts~\citep{gonzalez-marquez2024}, a common precursor of two different cell types in mass-spectroscopic data~\citep{li2018mass}, or continuous variation in single-cell transcriptomic data~\citep{scala2021phenotypic}.
Furthermore, embeddings can safeguard against errors and assist quality control because suspicious clusters, outliers, or processing errors become immediately apparent \citep{anscombe1973graphs,yanai2020hypothesis}. This helps to identify 
issues such as batch effects~\citep{polivcar2023embedding}, incorrect labels~\citep{diaz-papkovich_topological_2023}, or duplicated samples~\citep{bohm2022unsupervised}. 

Beyond hypothesis generation, embeddings can also support \emph{scientific communication}. 
When embeddings are used to present and explain research, they take the epistemic role of a \emph{map}. Just like a map provides a concise, abstracted overview of a territory, embeddings can facilitate data navigation or guide the narrative in communicating analysis results. 
Even more importantly, embeddings can increase the transparency of scientific communication. Even when the datasets are openly published alongside a manuscript, the readers will only rarely download and start exploring the data themselves.
Yet when presented with an adequate embedding, a reader can critically examine the structure revealed by the embedding, and this can help verify the results or prompt subsequent exploration and follow-up studies. 

In summary, we consider \textit{immediacy} and \textit{transparency} the two main virtues of two-dimensional embeddings. At the same time, the immediacy of vision is exactly what underlies the controversies and debates about embeddings: A visual impression can be misleading, yet hard to overcome (``seeing is believing''), calling into question the scientific objectivity of the whole enterprise. Indeed, critics have derisively compared two-dimensional embeddings to art \citep{chariSpeciousArtSinglecell2023} and to reading tea leaves, 
implying their unscientific or pseudoscientific character, and have called to forgo them entirely. 
Notably, two-dimensional embeddings have attracted more criticism than other EDA tools such as clustering, even though clustering also requires subjective choices \citep{kleinberg2002impossibility,luxburg2012clustering,hennig2015true} and can arguably introduce even larger distortions as it typically forces the data into a discrete (i.e., zero-dimensional) representation.
This criticism is largely due to the immediacy of vision leveraged by embeddings: While immediacy is a virtue, it comes with the risk of being deceiving. 

The discussion about the role of low-dimensional data representations in science has a long history, connecting to concerns regarding the use of visualizations in data analysis.  Already during the early beginnings of statistical computer graphics, \citet{anscombe1973graphs} opposed the idea that ``performing intricate calculations is virtuous, whereas actually looking at the data is cheating'' and defended the utility of data visualizations. And, as \citet{daston2007objectivity} show, the push against graphic aids to science extends back to at least the late nineteenth century. 
In this review, we adopt the stance that, when applied and interpreted with the appropriate care (see especially \cref{sec:best-practices}), low-dimensional embeddings constitute a useful tool that can add genuine scientific value by making the way science approaches big datasets less error-prone and more transparent.

\section{Overview of embedding methods}
\label{sec:algorithms}

\begin{table*}[t]
\caption{Selection of widely used embedding methods (Section~\ref{sec:algorithms}). Typical use cases and typical target dimensionality are based on the most common uses and not exhaustive. The `SVD' column indicates whether the exact solution can be obtained via singular value decomposition or eigendecomposition.}
\label{table:methods}
\centering
\begin{tabular}{ccccccc}
\toprule
Method & Loss function & Typical dim. & Parametric & SVD & Typical use cases\\
\midrule
PCA & Reconstruction error & 1--100 & linear & \checkmark & Preprocess; visualize \\
Metric MDS & Distance preservation & 2 or 3 & $\times$ & $\times$ & Visualize global structure \\
Isomap & Geodesic distance preserv. & 2 or 3 & $\times$ & \checkmark & Visualize contin.\ structures \\
Factor analysis & Likelihood & 1--10 & linear & $\times$ & Interpret latent variables \\
GTM & Likelihood & 1 or 2 & non-linear & $\times$ & Visualize contin.\ structures \\
Laplacian eig. & Constrained neighb.\ preserv. & 1--10 & $\times$ & \checkmark & Preprocess; visualize \\
LLE & Local reconstruction & 1--10 & $\times$ & \checkmark & Visualize contin.\ structures \\
PHATE & Potential distance preserv. & 2 or 3 & $\times$ & $\times$ & Visualize contin.\ structures \\
$t$-SNE & Neighbor preservation & 2 or 3 & $\times$ & $\times$ & Visualize clusters \\
UMAP & Neighbor preservation & 2--10 & $\times$ & $\times$ & Visualize clusters; preprocess \\
Autoencoders & Reconstruction error & 2--100 & non-linear & $\times$ & Preprocess; visualize \\
\bottomrule
\end{tabular}
\end{table*}

Given a collection of high-dimensional data samples ${\mathbf x \in \R^D}$, 
dimensionality-reduction techniques construct low-dimensional representatives $\mathbf z \in \R^d$ with $d\ll D$, 
aiming to preserve certain characteristics of the original data. The low-dimensional samples 
can be obtained via a function $f_W\colon\R^D\to\R^d$ parametrized by a set of learnable parameters $W$, or in a non-parametric way when only the embedding positions themselves are learned.
Representing high-dimensional data in a lower-dimensional space
usually means sacrificing some information, due to the loss of degrees of freedom and limited expressiveness of the restricted space. 

The choices of which information to discard and which characteristics to retain form the foundation of any dimensionality-reduction strategy. 
Today, myriad methods exist, aiming to preserve different characteristics (e.g., variance or pairwise distances), with different mapping strategies (parametric or non-parametric, linear or nonlinear), and using diverse optimization approaches. 
Multiple books have been written on the subject \citep{lee2007nonlinear, lespinats2022nonlinear, ghojogh2023elements}. 
In this section, we give a non-exhaustive, concise overview of existing methods, their relationships, and the trade-offs they involve, focusing on techniques that are historically important or currently commonly used in practice (\Cref{table:methods}).

\subsection{Linear methods}

\begin{figure}[!t]
\centering
\includegraphics[width=\linewidth]{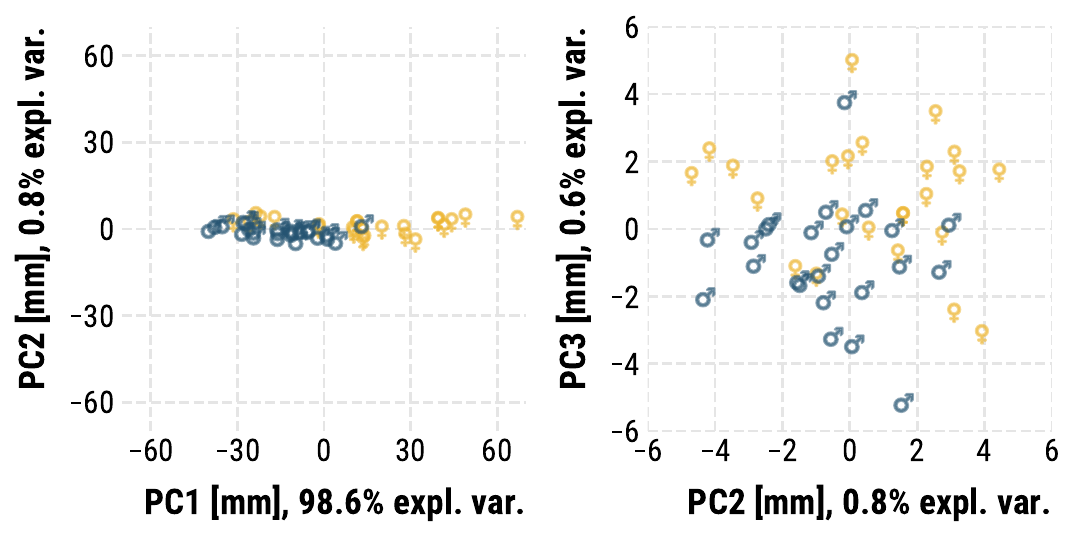}
\caption{PCA of turtle carapace measurements ($n=48$ specimens) from \citet{jolicoeur_size_1960}, the first paper to use PCA for 2D visualization. In this species, females are larger than males.}
\label{fig:turtles}
\end{figure}

Developed in the early 20th century as one of the very first dimensionality-reduction algorithms, principal component analysis (PCA) \citep{pearson1901liii, hotelling1933analysis} is defined as a linear projection from the high-dimensional data space onto orthogonal axes, yielding uncorrelated variables known as principal components (PCs). PCs are ordered by decreasing variance, with each PC capturing a certain fraction of the total original variance. Among all linear mappings to a given target dimension, the PCA projection maximally preserves data variance and minimizes the reconstruction error. The exact PCA solution can be obtained in terms of the eigendecomposition of the data covariance matrix.

PCA has countless applications \citep{jolliffe1986principal, ringner2008principal} both as a data-preprocessing method and as a data-visualization method. 
When PCA is used for data preprocessing, the data is often reduced to 10--100 dimensions for downstream computational analysis. 
In contrast, when PCA is used for data visualization, the leading 2--3 PCs are plotted directly.
The orthogonal linear projection and its intuitive objective function make PCA visualizations particularly easy to interpret. The earliest visualization of a 2D PCA embedding that we identified is a 1960 paper \citep{jolicoeur_size_1960} analyzing turtle carapace measurements (Figure~\ref{fig:turtles}). In the 1970s, related methods were developed specifically for visualization, such as the biplot \citep{gabriel1971biplot} and correspondence analysis \citep{greenacre1984theory}. The simplicity and popularity of PCA motivated the development of numerous extensions, such as sparse PCA \citep{zou2006sparse}, robust PCA \citep{candes2011robust}, kernel PCA \citep{scholkopf1997kernel}, and probabilistic PCA \citep{tipping1999probabilistic}.

Similarly defined as a linear transformation, independent component analysis (ICA) \citep{comon1994independent} generalizes PCA by looking for \emph{independent} instead of \emph{uncorrelated} components. 
While not as popular as PCA, ICA has been widely used in some fields, e.g., for blind source separation in electric brain recordings \citep{makeig1995independent}.

\subsection{Distance-preserving methods}
\label{sec:distance_preservation}

While linear methods construct an explicit mapping from the high-dimensional to the low-dimensional space, some other methods directly optimize the placement of low-dimensional points. Such methods are sometimes imprecisely referred to as \textit{nonlinear} dimensionality reduction, but it is more accurate to call them \textit{non-parametric}. One prominent example dating back to the 1950s and 1960s is multidimensional scaling (MDS), which maximizes the preservation of high-dimensional distances in the low-dimensional embedding space \citep{borg1997modern}. While traditional implementations are slow, the recently introduced SQuadMDS algorithm \citep{lambert2022squadmds} provides a fast approximation suitable for large datasets.

MDS exists in several flavors. What is described above is called \emph{metric MDS}. There is a simpler variant known as \emph{classical MDS}, or Torgerson MDS \citep{torgerson1952multidimensional}, or principal coordinates analysis (PCoA) \citep{gower1966some}, which can be solved exactly via eigendecomposition and, when applied to pairwise \emph{Euclidean} distances, is equivalent to PCA. \emph{Non-metric MDS} additionally optimizes a monotonic transformation of high-dimensional distances \citep{shepard1962analysis, shepard1962analysisII, kruskal1964multidimensional, kruskal1964nonmetric}, but scalable implementations are lacking. Weighted distance-preservation schemes include Sammon's mapping \citep{sammon1969nonlinear} and curvilinear component analysis (CCA) \citep{demartines1997curvilinear}, which favor the preservation of short high-dimensional and low-dimensional distances, respectively.

\subsection{Probabilistic methods}

Another class of methods is based on postulating a \emph{generative} model: Probabilistic methods represent the data using a probabilistic mapping from some unobserved \emph{latent} variables to the high-dimensional space. 
Then, the expectation-maximization (EM) algorithm is employed to find the most likely mapping and the corresponding latent variables.
This approach is motivated by the idea that a small number of latent variables drives the meaningful variability in the high-dimensional data. 
The simplest example is the probabilistic formulation of PCA~\citep{tipping1999probabilistic}, which showed that if the latent variables are continuous, the mapping from the latent space to the data space is linear, and the high-dimensional noise is equally strong in all dimensions, then the latent variables are given by the principal components. 
The probabilistic perspective on PCA allows for many generalizations, such as mixtures~\citep{tipping1999mixtures}, hierarchies~\citep{bishop1998hierarchical}, and Bayesian formulations~\citep{bishop1998bayesian}. 

When allowing the high-dimensional noise to have different magnitudes in different dimensions, the same model leads to factor analysis. This has been developed largely in parallel to PCA in the early 20th century \citep{spearman1961general, thurstone1931multiple}, and it has been popular in psychometrics, usually as a tool to extract interpretable latent factors and not for visualization~\citep{everett1984latent}. 
The linear map expresses the \emph{factor loadings}, while the diagonal elements represent independent noise variances for each variable.  Extensions of factor analysis to time-series data have been developed in neuroscience~\citep{yu2008gaussian}.

Probabilistic PCA and factor analysis are limited by their linearity and the restricted form of the latent-variable distribution. 
Increasing their expressive power brings computational challenges: arbitrary non-linear functions and distributions usually lead to intractable algorithms. 
Generative topographic mapping (GTM)~\citep{bishop1998gtm} hence defines the latent distribution as a finite regular grid of delta functions over the low-dimensional (usually 2D) latent space and uses non-linear mapping functions. 
The stretching and magnification of the latent-space representation to fit the data can be measured using differential geometry; the separation between clusters can be visualized by plotting the magnification factors in latent space~\citep{bishop1997magnification}. 
Similarly, the curvature of the manifold can be used to quantify how well the manifold fits the data~\citep{tino2001using}. 
Extensions of GTM can model discrete variables and heterogeneous data~\citep{nabney2005semisupervised}, as well as time series by using a hidden Markov model~\citep{bishop1997gtm}. More recently, the restriction to a grid of delta functions has been relaxed through the use of Gaussian process mappings~\citep{lawrence2003gaussian}, which tends to provide a smoother distribution of points in the latent space.

\subsection{Spectral methods} 
\label{sec:spectral}

This group of methods is based on the \emph{manifold assumption}, i.e.,  the hypothesis that the original data points lie on a low-dimensional manifold within the high-dimensional space. Spectral methods approximate the data manifold by representing the neighbor relations between data points as a graph\,---\,e.g., in a $k$-nearest-neighbor ($k$NN) graph, the data points (nodes) are connected (edges) to their $k$ nearest neighbors in the high-dimensional space. 
The core idea behind spectral methods is to find a low-dimensional embedding of the data that preserves the structural relationships defined by this graph. Solving the resulting objectives usually involves computing the \emph{spectral} decomposition (eigendecomposition) of matrices derived from the weighted adjacency matrix of the graph. Most notably, Laplacian eigenmaps~\citep{belkin2002laplacian} perform an eigendecomposition of the normalized graph Laplacian, which yields a constrained embedding such that neighboring points in the graph end up close together in the embedding.
Theoretical results show that Laplacian-based embeddings approximate the geometry of the underlying data manifold~\citep{belkin2006convergence,singer2006graph}.

Closely related to Laplacian eigenmaps are diffusion maps~\citep{coifman2006diffusion, nadler2006diffusion}. Here, the \textit{diffusion distance} between any two points is defined based on the similarity of fixed-length random walks on the graph emanating from these two points. Diffusion maps approximate this distance with an embedding given by scaled eigenvectors of the normalized random-walk graph Laplacian.
Specific normalizations of this matrix can yield embeddings that are invariant to the density of the sampled data points~\citep{coifman2006diffusion}, which is important when non-uniform sampling is an artifact of the data-acquisition process, rather than a reflection of the geometry of the data~\citep{lafon2006data}.
Extensions include considering information on the local curvature of the high-dimensional point cloud~\citep{singer2012vector, singer2017spectral} or approximating the commute-time distances that consider random walks of arbitrary length between pairs of points~\citep{taylor2012random}. 

Other methods compute the eigendecomposition of more specialized data-dependent matrices arising from the $k$NN graph. Locally linear embedding (LLE)~\citep{roweis2000nonlinear} aims to preserve local relationships by reconstructing each data point as a linear combination of its neighbors, with extensions capturing higher-order information~\citep{donoho2003hessian}. In contrast, maximum variance unfolding (MVU) \citep{weinberger2006introduction, song2007colored} aims to maximize variance in the low-dimensional space while maintaining the local distances defined by the nearest-neighbor graph.

Similar to PCA, eigendecompositions in spectral methods yield a whole sequence of eigenvectors, and more than two eigenvectors may often be needed to adequately represent the graph structure. 
As an alternative, to obtain a 2D or 3D embedding that reflects distances on the graph, MDS (\cref{sec:distance_preservation}) can be applied to pairwise graph-based distances.
This was first suggested in Isomap, which amounts to classical MDS of the geodesic distances on the $k$NN graph \citep{tenenbaum2000global}. Isomap works well for revealing smooth nonlinear manifolds in low-noise settings, but it can suffer from high-dimensional noise because geodesic distances can be very sensitive to shortcut edges in the graph. A more recent approach called PHATE~\citep{moon2019visualizing} uses metric MDS of potential distances, which are a variant of diffusion distances inspired by information geometry. Both are less influenced by shortcuts, making PHATE more robust than Isomap.
Variants of PHATE have been proposed for multiscale embeddings~\citep{kuchroo2022multiscale} and for capturing the dynamics of evolving systems~\citep{gigante2019visualizing}. 

Most spectral approaches obtain a \emph{global} embedding of the data via an eigendecomposition of a graph matrix, e.g., the graph Laplacian. 
In contrast, bottom-up manifold-learning approaches such as local tangent space alignment (LTSA) \citep{zhang2004}, low distortion local eigenmaps (LDLE)~\citep{kohli2021ldle}, and Riemannian alignment of tangent spaces (RATS)~\citep{kohli2024rats}, first construct separate low-dimensional embeddings of local neighborhoods in the data (using local PCA or subsets of the eigenvectors of the graph Laplacian) and then align them to obtain a single global embedding.

\subsection{Neighbor-embedding methods}
\label{sec:neighbor_embeddings}

\begin{figure}[t]
    \centering
    \includegraphics[width=\linewidth]{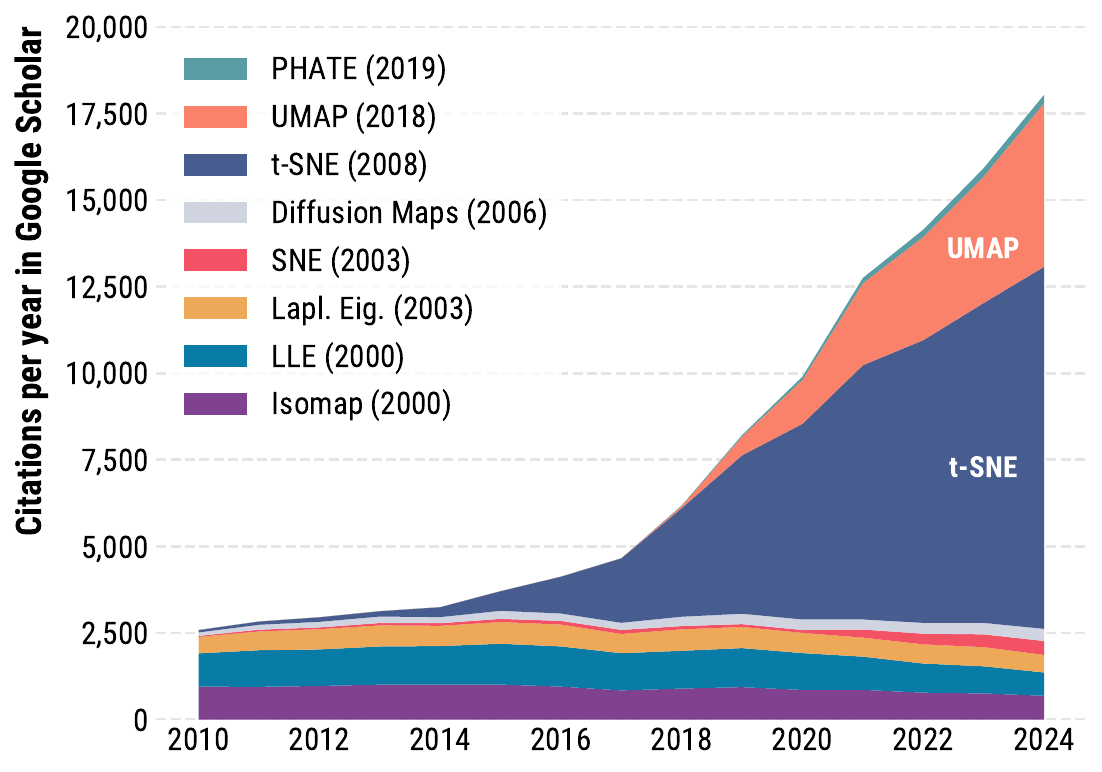}
    \caption{Yearly citations (according to Google Scholar) of some highly cited papers introducing dimensionality-reduction methods. Older methods (such as PCA or MDS) are not shown due to the lack of a single canonical citation.}
    \label{fig:citations}
\end{figure}

Neighbor-embedding methods, going back to the seminal stochastic neighbor embedding (SNE) algorithm \citep{hinton2002stochastic}, are also based on neighbor graphs of the data, such as $k$NN graphs. 
These methods aim to construct an embedding such that high-dimensional neighbors remain neighbors in the embedding while high-dimensional non-neighbors are placed far apart. 
While metric MDS aims to preserve all pairwise high-dimensional distances, large and small, SNE only aims to preserve nearest neighbors, corresponding to the smallest pairwise distances. 

The focus on nearest-neighbor preservation over global distance preservation offers one main advantage:
MDS often fails to generate useful embeddings because high-dimensional distances tend to be all similar to each other (a phenomenon known as norm concentration \citep{aggarwal2001surprising,francois2007concentration}), one manifestation of the curse of dimensionality. 
Hence, these high-dimensional distances cannot be meaningfully reproduced in 2D (see \cref{sec:showcasing}). 
Neighbor embeddings tend to work better than MDS because they focus on preserving neighbors, irrespective of what the distances between them originally were \citep{lee2011shift}. 
Consequently, methods based on neighbor embeddings have gained traction in recent years, as evidenced by the growing popularity of their most prominent representatives, $t$-SNE and UMAP (\Cref{fig:citations}).

A modification of the original SNE, $t$-SNE \citep{maaten2008visualizing} allows some of the neighbors to drift further away. As a result, it is less influenced by the shortcut $k$NN edges (see also \cref{sec:spectral}) and can often produce a better overall embedding without points crowding together too strongly.
Motivated by the practical success of the SNE framework, many modifications have been proposed, such as hierarchical~\citep{pezzotti2016hierarchical}, multiscale~\citep{lee2015multi, cdb2022fastms}, and heavy-tailed~\citep{yang2009heavy,kobak2019heavy} SNE variants.
The growing interest in visualizing datasets with ever-increasing sample size has led to the development of accelerated $t$-SNE implementations that use approximate $k$NN graphs and physics-inspired methods for approximate optimization~\citep{yang2013scalable, van2014accelerating, vladymyrov2014linear, linderman2019fast, pezzotti2019gpgpu, artemenkov2020ncvis}. Modern implementations~\citep{polivcar2019opentsne} easily scale to tens of millions of samples. 

UMAP \citep{mcinnes2018umap,healy2024uniform} is based on an earlier accelerated modification of $t$-SNE~\citep{tang2016visualizing}. The main difference between UMAP and $t$-SNE can be understood when interpreting neighbor-embedding algorithms as physical systems, where points move in the embedding space under attractive forces between neighbors and indiscriminate repulsive forces between all points until an equilibrium is reached~\citep{carreira2010elastic}. UMAP employs stronger attraction forces than $t$-SNE \citep{damrich2023from}, resulting in more compact and pronounced clusters, but both methods belong to a whole attraction-repulsion spectrum of neighbor embeddings \citep{bohm2022attraction}. Other differences between UMAP and $t$-SNE, such as the choice of edge weights, have been shown to be of little importance \citep{damrich2021umap}. UMAP's popularity inspired the development of multiple related methods \citep{amid2019trimap, wang2021understanding}. Recently, UMAP embeddings to 5--10 dimensions have been used for density clustering in some application areas \citep{grootendorst2022bertopic,diaz-papkovich_topological_2023,healy2024uniform}. We believe more work is needed to better understand and validate this clustering approach (see \cref{sec:challenges}).

Similar to MDS, optimizing a neighbor-embedding objective is non-convex. Hence, different initializations can result in different embeddings, corresponding to different local optima. Modern implementations benefit from informative initializations \citep{kobak2021initialization}, and various optimization heuristics such as dimensionality annealing \citep{vladymyrov2019no} have been suggested. 

Neighbor-embedding methods can also be interpreted as low-dimensional layouts of $k$NN graphs. Other graph-layout algorithms, such as force-directed graph layouts, can be used on $k$NN graphs, also resulting in neighbor embeddings. One example is ForceAtlas2~\citep{jacomy2014forceatlas2}, which leads to even stronger attractive forces compared to UMAP~\citep{bohm2022attraction}. Conversely, UMAP and $t$-SNE can be used as graph layouts for generic graphs, such as citation networks \citep{bohm2025node}.

\subsection{Parametric methods}

Except for linear and probabilistic methods, most methods described above are non-parametric, i.e., they construct an embedding without learning an explicit mapping between the high-dimensional and the low-dimensional coordinates. While flexible, one limitation of these kinds of approaches is that adding out-of-sample data into existing embeddings is not straightforward \citep{bengio2004learning}. One family of non-linear \emph{parametric} methods are auto-encoders \citep{hinton2006reducing}, where geometric or topological constraints on the non-linear mapping have been used to obtain more geometrically accurate parametric embeddings of the data \citep{jia2015laplacian,li2020variational,li2021invertible,peterfreund2020local,moor2020tae, duque2023grae,nazari2023geometric}.

Many non-parametric methods can be converted into their parametric variants using various neural network architectures. However, the effects of such parametrization have only recently been systematically explored~\citep{duque2023grae,huang2024navigating}. Many such approaches have been proposed for spectral embeddings \citet{mishne2019diffusion,shaham2018spectralnet,pai2019dimal,duque2020extendable} and  
neighbor-embedding methods \citep{maaten2009learning, bunte2012general, gisbrecht2015parametric, sainburg2021parametric, carreira2015fast, damrich2023from}.

\subsection{Supervised methods}

All methods surveyed above are \textit{unsupervised}: They do not use class labels that may be available in the dataset and focus on the faithful preservation of structure in the high-dimensional data. In contrast, \textit{supervised} dimensionality reduction looks for an embedding with high separation of predefined classes. These two goals may be conflicting if the most prominent high-dimensional structure is not driven by the class labels. 

A supervised linear method called linear discriminant analysis (LDA) \citep{rao1948utilization} can be seen as a supervised version of PCA, seeking to maximize the separation between means of projected classes while minimizing variance within each projected class. Two-dimensional LDA scatter plots were in use even before PCA scatter plots \citep{rao1948utilization,jolicoeur1959multivariate}. Related methods include demixed PCA \citep{kobak2016demixed} for data with multiple sets of class labels, as well as canonical correlation analysis \citep{hotelling1936relations} and reduced-rank regression \citep{izenman1975reduced} for data with multiple continuous labels. 

Another linear method, neighborhood component analysis \citep{goldberger2004neighbourhood} finds a projection that maximizes the $k$NN classification accuracy in the embedding. Other works use the class information to construct supervised pairwise distances and then feed them into unsupervised embedding methods \citep{venna2010information, bunte2012limited}\,---\,supervised variants of Isomap \citep{geng2005supervised, li2006supervised} and PHATE \citep{rhodes2021random} were developed following this principle. Similarly, class information can also be used to inform neighbor-embedding algorithms~\citep{yu2017deep, hajderanj2019new, cheng2021supervised}.

\section{Example applications}
\label{sec:showcasing}

Representing high-dimensional data in low-dimensional spaces requires trade-offs, and each method will preserve different aspects of the input data to a different extent (\cref{sec:algorithms}). Therefore, the utility of a given method will strongly depend on the data type and on the analysis goal. To illustrate this, we apply six popular embedding methods\,---\,PCA, MDS, Laplacian Eigenmaps (LE), PHATE, $t$-SNE, and UMAP\,---\,to create 2D visualizations of real-world datasets from three data modalities and scientific domains: text data, single-cell transcriptomics data, and population-genetics data.
We also quantitatively evaluate the different methods regarding their capacity to preserve local and global structure.  

\paragraph{Text data}

\begin{figure*}[t]
\centering
\includegraphics[width=\textwidth]{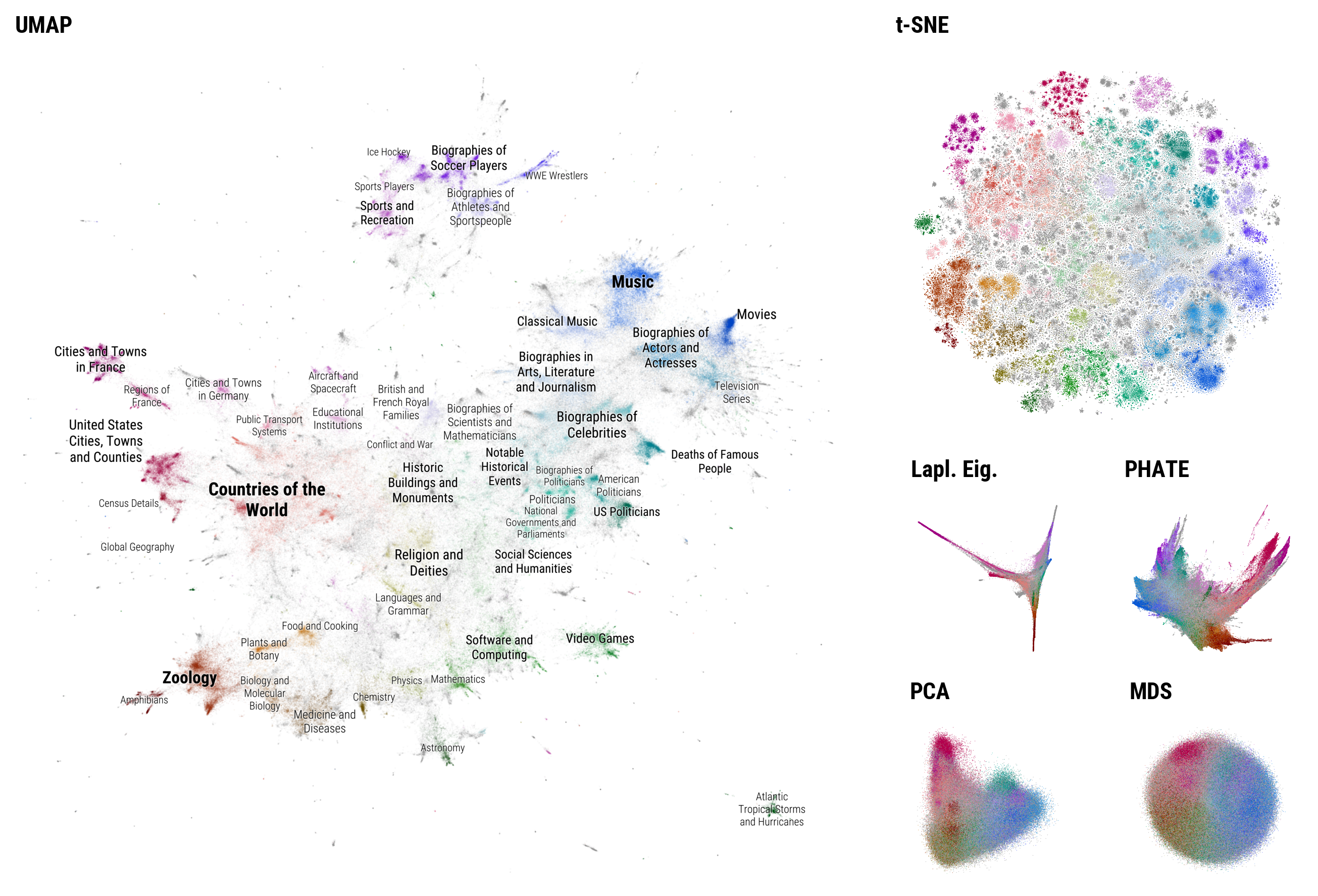}
\caption{2D embeddings of 485\,900 text paragraphs from the Simple English Wikipedia, processed using an LLM. Colors correspond to clusters identified by HDBSCAN in 2D UMAP embedding, which were automatically labeled using a generative LLM. These clusters only serve to annotate the structure visible in the UMAP embedding, and should not be used in downstream analysis (see \Cref{sec:best-practices}). The first two principal components together explained 6.6\% of the total variance.
}
\label{fig:showcasing_wikipedia}
\end{figure*}

Neighbor embeddings have been used in recent years to visualize and obtain explorable embeddings of large document corpora, such as library contents \citep{schmidt2018stable}, collections of philosophy papers \citep{noichl_modeling_2021}, or biomedical articles \citep{gonzalez-marquez2024}. They can also be used to investigate more specialized textual formats, like mathematical formulas \citep{noichlHowLocalizedAre2023a}. Here, we visualized a collection of 485\,900 text paragraphs taken from the Simple English Wikipedia and represented as 768-dimensional vectors using a language model (see Methods) (\Cref{fig:showcasing_wikipedia}). Both UMAP and $t$-SNE embeddings exhibited numerous clusters that were semantically meaningful and corresponded to well-defined topics, such as \textit{video games} or \textit{mathematics}. While the UMAP embedding showed more clearly separated larger clusters, $t$-SNE emphasized smaller clusters corresponding to sub-topics, for example, individual countries within the \textit{world countries} topic. 

Manifold-learning methods such as PHATE and Laplacian eigenmaps produced embeddings that were more difficult to interpret, 
likely because the input data did not have prominent continuous manifold structures. 
Finally, PCA and especially MDS yielded fuzzy 2D embeddings lacking useful structure. Indeed, MDS aims to preserve high-dimensional pairwise distances, which in this case tend to be all similar and hence cannot be meaningfully reproduced in 2D. Note that PCA and Laplacian eigenmaps yield a sequence of eigenvectors, and more than two dimensions are needed to represent the data accurately. In this dataset, the first two principal components represented less than 7\% of the total variance.

\paragraph{Single-cell transcriptomics data}

\begin{figure*}[t]
\centering
\includegraphics[width=\textwidth]{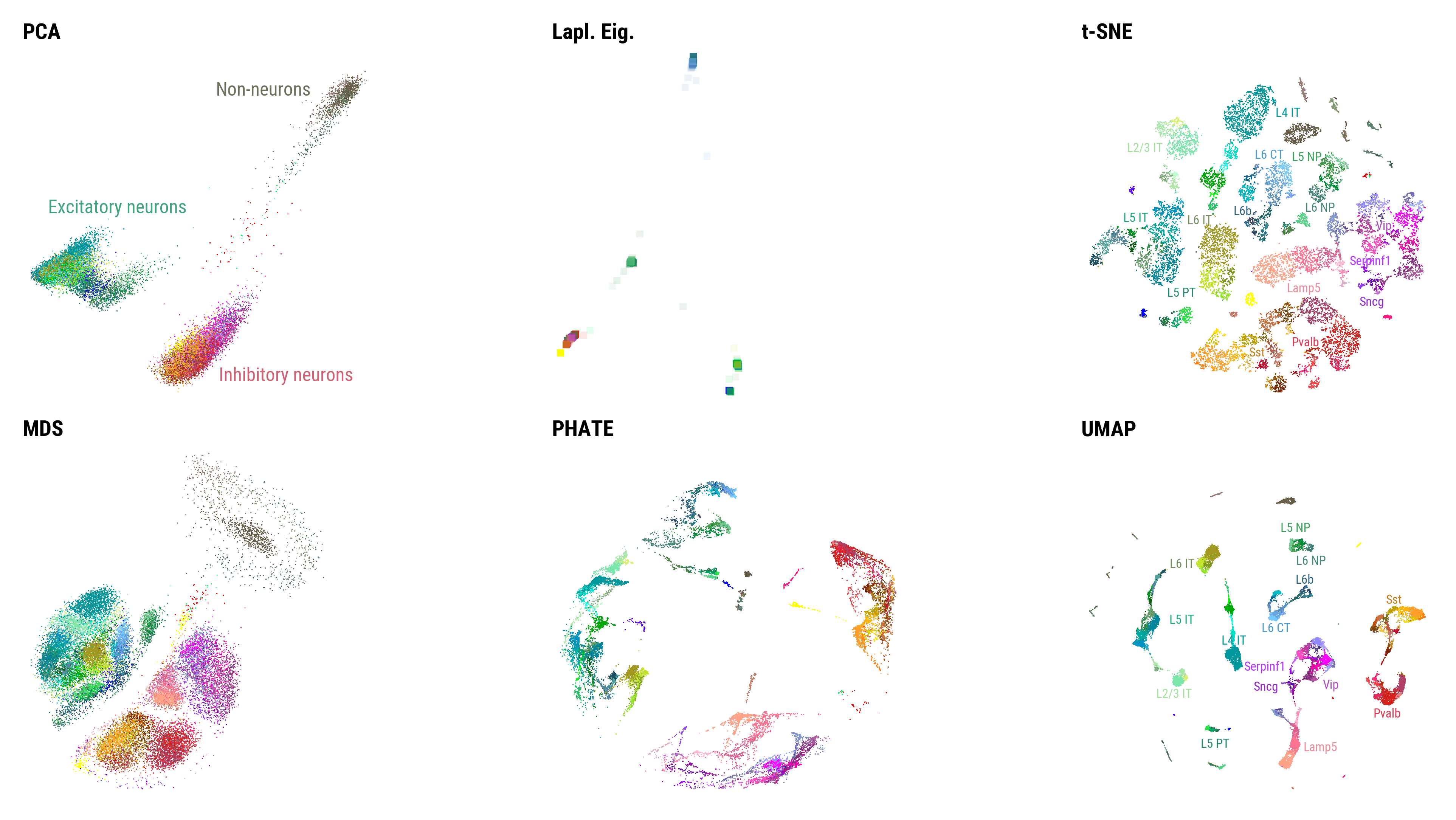} 
\caption{2D embeddings of 23\,800 cells from the mouse cortex \citep{tasicSharedDistinctTranscriptomic2018}. Colors correspond to transcriptomic cell types, taken from the original publication. The first two principal components explained 49.1\% of the variance of the preprocessed data. As Laplacian eigenmaps had many almost-overlapping points, they are shown with larger semi-transparent markers.}
\label{fig:showcasing_tasic}
\end{figure*}

\begin{figure*}[t]
\centering
\includegraphics[width=\textwidth]{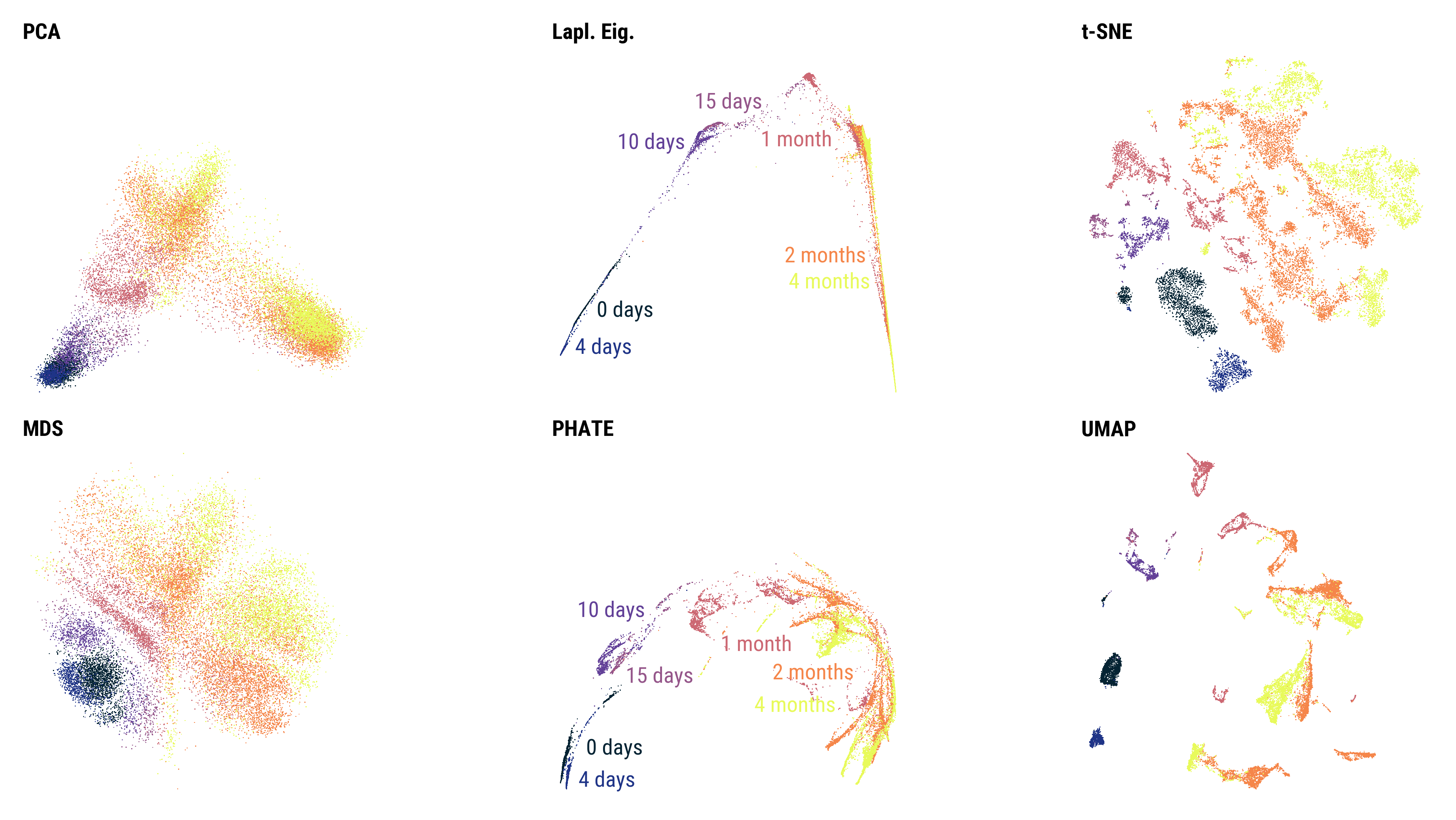} 
\caption{2D embeddings of 20\,300 cells from primate brain organoids \citep{kantonOrganoidSinglecellGenomic2019}. Colors correspond to sample age (from 0 to 120 days). The first two principal components explained 48.9\% of the variance of the preprocessed data.}
\label{fig:showcasing_kanton}
\end{figure*}

In the field of single-cell biology, two-dimensional visualizations, typically based on neighbor embeddings or manifold-learning methods, have become a commonplace tool: they assist with the exploration of cell types and their development and provide a concise visual data summary in publications. 
Here, we use two datasets for which biology dictates very different data geometry. One dataset \citep{tasicSharedDistinctTranscriptomic2018} contains 23\,800 cells from the adult mouse brain and has a large number of hierarchically organized cell types (\Cref{fig:showcasing_tasic}). The other dataset \citep{kantonOrganoidSinglecellGenomic2019} contains 20\,300 cells from primate brain organoids collected during organoid development and has prominent one-dimensional organization corresponding to developmental time (\Cref{fig:showcasing_kanton}). We use standard preprocessing steps (see Methods), including dimensionality reduction from tens of thousands of original features (genes) to 50 principal components. 

In the dataset from \citet{tasicSharedDistinctTranscriptomic2018}, PCA and MDS embeddings clearly emphasized three major classes of cells: inhibitory neurons, excitatory neurons, and non-neural cells (\Cref{fig:showcasing_tasic}), with MDS and PHATE indicating further variability within some of the classes. The UMAP and $t$-SNE embeddings suggested that cells were grouped into around a dozen well-separated cell families (with UMAP compressing them more than $t$-SNE), but they did not convey any information on the large-scale organization of such families into classes. To some extent, this can be remedied by multiscale $t$-SNE that combines local quality of $t$-SNE with the global layout similar to MDS \citep{cdb2022fastms}. Notably, Laplacian eigenmaps embedding collapsed major classes almost to single points, corresponding to disconnected components in the $k$NN graph. While this property of Laplacian eigenmaps is useful for downstream clustering known as spectral clustering \citep{ng2001spectral,shi2000normalized}, it arguably counteracts meaningful visualization.

The appeal of manifold-learning methods can be seen when embedding the dataset from \citet{kantonOrganoidSinglecellGenomic2019} (Figure~\ref{fig:showcasing_kanton}). Here, the first component of  Laplacian eigenmaps corresponded to developmental time. The same developmental trajectory was visible in the PHATE embedding (as well as in the ForceAtlas2 layout of the $k$NN graph of the data, see \citet{bohm2022attraction}). In contrast, UMAP and especially $t$-SNE embeddings highlighted individual clusters and did not show the continuous developmental structure. 
These observations illustrate why diffusion-based methods, emphasizing continuous variation, are often used in developmental single-cell biology~\citep{haghverdi2015diffusion, angerer2016destiny, moon2019visualizing}. Using the same dataset, \citet{damrich2024visualizing} argued that adjusting the strength of attractive forces in neighbor embedding algorithms can be helpful when working with developmental data.  

\paragraph{Population-genetics data}

\begin{figure*}[t]
\centering
\includegraphics[width=\textwidth]{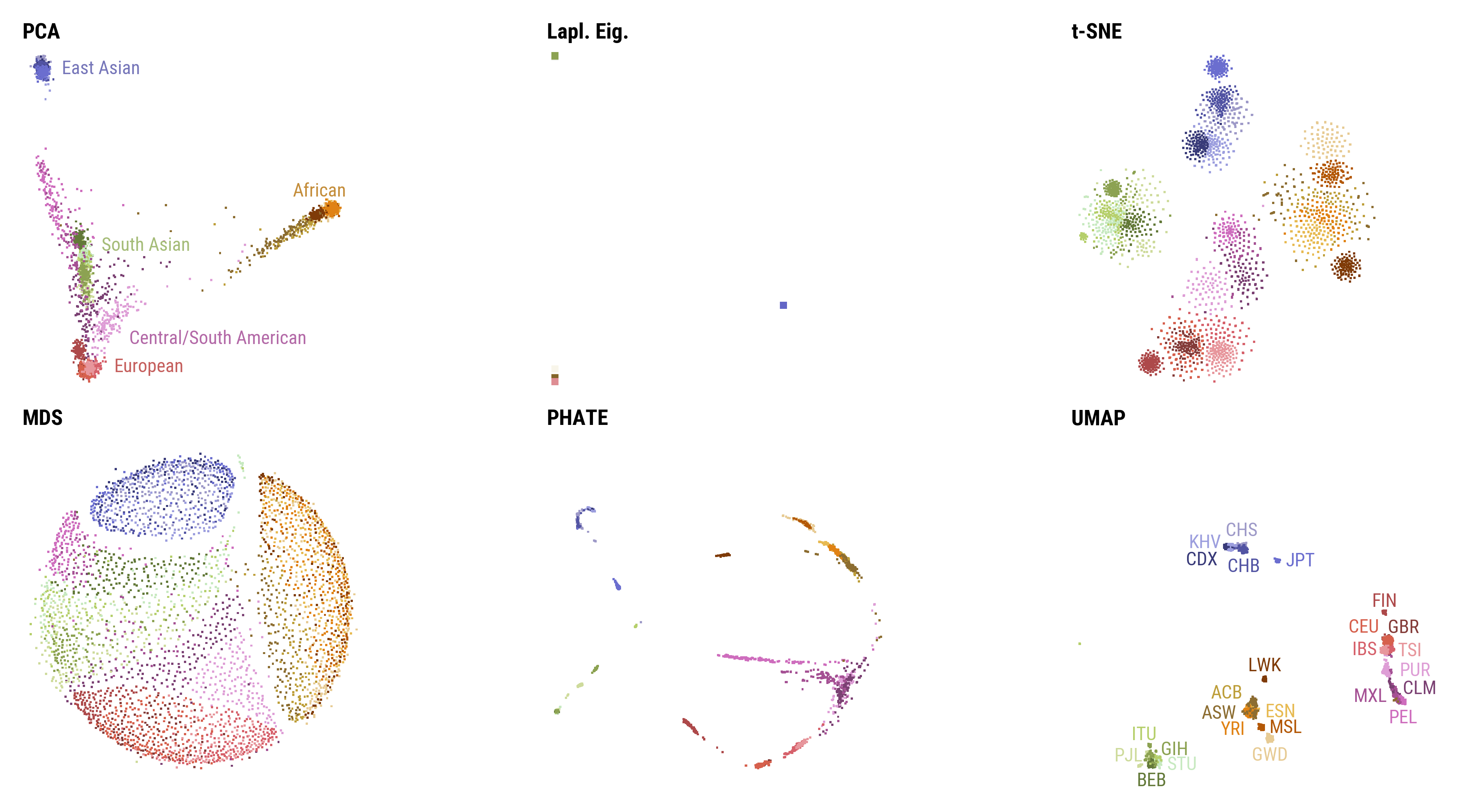} 
\caption{2D embeddings of 3\,450 human genotypes from $26$ global populations \citep{global_2015}. Colors represent the sampling population. The first two principal components together explained 5.8\% of the total variance. For population abbreviations used to annotate the UMAP embedding, see the original publication. As Laplacian eigenmaps had many almost-overlapping points, they are shown with large semi-transparent markers. 
}
\label{fig:showcasing_genomes}
\end{figure*}

Population-genetic studies use 2D embeddings to visualize population structure that arises from patterns of non-random mating over generations, providing insights into demographic history. 
We used 3\,450 human genotypes from the 1000 Genomes Project \citep{global_2015}, which sampled data from 26 populations around the world (\Cref{fig:showcasing_genomes}). We employed standard preprocessing to filter data to 54\,000 single-nucleotide polymorphisms (SNPs) as features (see Methods).
UMAP and $t$-SNE embeddings formed clusters of individuals sharing recent genetic ancestry and hence represented the 26 sampled populations. UMAP formed tighter and larger clusters (e.g., East Asian, European, African, etc.), whereas $t$-SNE exhibited smaller clusters, including multiple clusters consisting of just several genotypes and corresponding to single families (siblings and parents). However, individuals who share ancestry from multiple clusters (e.g., a child with parents from different continents) may appear in such embeddings within one cluster rather than between them, i.e., these methods can exaggerate the separation between clusters.

In contrast, PCA showed more continuous structure with several prominent axes (Figure~\ref{fig:showcasing_genomes}). Since the largest source of genetic variation in this data is geographic distance between populations, the terminals of the PC axes represented the geographical ancestry regions: Africa, South Asia, East Asia, and Europe. Central and South American individuals, who tend to have recent ancestry from Europe and Africa in addition to their own indigenous ancestry, appeared between these clusters.
However, PCA required more than two components to fully represent this structure, making the 2D figure difficult to read correctly and suggesting that visualizing further PCs can be useful (for example, the green points overlapped with other points in 2D, but were separated along PC3). Notably, MDS embedding of this dataset produced a disc with weakly separated classes: Since the leading PCs captured only a small fraction of the total variance (5.8\% by the first two PCs in contrast to nearly 50\% in our single-cell transcriptomics examples), the high-dimensional pairwise distances were dominated by other sources of variation.

\paragraph{Quantitative evaluation}

To quantitatively compare the showcased embedding methods, we computed several measures that capture how well an embedding preserves global and local structure (\Cref{table:qa}). 
Across the four datasets, $t$-SNE embeddings had the highest local quality, while PCA and MDS embeddings had the highest global quality (\Cref{fig:metrics}),  
in line with existing single-cell benchmarks \citep{huang2022towards, lause2024art, sun2019accuracy, wang2023comparative,  xiang2021comparison}. 
For the data from \citet{kantonOrganoidSinglecellGenomic2019} with its underlying continuous structure, PHATE and Laplacian Eigenmaps also showed higher global quality than $t$-SNE and UMAP.
Importantly, these metrics do not quantify all relevant aspects of an embedding (see \Cref{sec:challenges}). 

Overall, across the four datasets, we demonstrated that different embedding methods represent different aspects of the high-dimensional data.
Practitioners need to be aware of the trade-offs involved and emphasize them in their scientific communication, in particular when creating visualizations\,---\,such as those of human genotype data\,---\,that can potentially fuel societally contentious debates. 
In the next section, we provide some guidance toward this end.

\section{Guidance and best practices}
\label{sec:best-practices}

Low-dimensional embeddings are tools to examine and represent data\,---\,analogous to how microscopes are used to inspect cells and maps are designed to chart territories.
Working with each tool requires taking multiple careful decisions, 
such as how to calibrate a microscope or what projection to use for a map. 
The choices made to create low-dimensional embeddings should be equally deliberate.

In this section, we have compiled some of the best practices for working with low-dimensional embeddings based on our experience.
We group our guidance into four categories: preparation, exploration, presentation, and communication.
Our exposition complements the work by \citet{nguyen2019ten}, which mainly focuses on best practices in the context of linear dimensionality-reduction techniques. 
We provide recommendations regarding a broader range of dimensionality-reduction methods, with an emphasis on visualization.

\paragraph{Preparation}

Here, we discuss how to prepare the data and choose the appropriate embedding method.

\begin{enumerate}
    \item \textbf{Choose the embedding approach.} Are you interested in a 2D or 3D embedding for visualization purposes or in higher-dimensional embeddings for downstream analysis (e.g., classification, clustering, or topological analysis)? Is interpretability of individual dimensions important? What aspects of the data are of primary interest: fine-grained clusters? large-scale clusters? continuous manifolds? These questions should guide the choice of method, as discussed in \Cref{sec:algorithms,sec:showcasing}. 

    \item \textbf{Preprocess the data and select features.}
    Different types of data may require different preprocessing strategies: sample selection based on quality control, feature selection based on variability, normalization of samples, standardization of features, log-transformation of values, etc. All of this may substantially alter the data distribution and, hence, the embedding results. Note that changing the measurement units of individual features (e.g., from meters to millimeters) can strongly affect the data geometry as well. The optimal preprocessing workflow is often field-specific and benefits from familiarity with the data-generation process. 
    In some cases, features can be constructed using pretrained models\,---\,for example, texts, images, or audio samples can be converted into high-dimensional vectors (often also called embeddings) using a pretrained language model or a convolutional neural network. 
    However, one should be aware that this strategy also introduces any biases from these models into the preprocessed data.

    \item \textbf{Pick an appropriate distance or similarity measure.}
    Most embedding methods rely on a distance or similarity measure between data points. The default distance is typically Euclidean, but other distances, e.g., cosine, Mahalanobis, or Wasserstein, can be used instead, based on the data type or expected local geometry~\citep{talmon2013empirical,mishne2016hierarchical,otoole2023novelty,benisty2024rapid}. 
    Domain knowledge can also determine more effective domain-specific similarity measures that will impact the quality of an embedding~\citep{lozupone2005unifrac,talmon2012single}.
    
    \item \textbf{Determine the number of PCs to keep.}
    For many types of data, standard preprocessing pipelines include a denoising step with PCA, preserving only a subset of PCs as input to a secondary embedding algorithm. This step can affect the results, especially if a large fraction of variance is discarded in the PCA step. See \citet[Sec 6.1]{jolliffe1986principal} for strategies to choose the number of PCs. 
\end{enumerate}

\paragraph{Exploration}

Here, we summarize what to keep in mind when exploring an embedding of a given dataset.

\begin{enumerate}
    \setcounter{enumi}{4}
    
    \item \textbf{Look at multiple embedding methods.}\label{item:multiple-embeddings}
    Visualizing data with multiple embedding methods is analogous to using multiple types of microscopes on a biological sample. In \Cref{sec:showcasing}, we showed how PCA, MDS, UMAP, $t$-SNE, LE, and PHATE can highlight different aspects of the data. However, to be able to interpret the differences meaningfully, it is important to be aware of the differences between the algorithms and the trade-offs involved (\Cref{sec:algorithms,sec:showcasing}). 

    \item \textbf{Consider hyperparameters.}\label{item:consider-hyperparameters}
    Many methods have \emph{conceptual} hyperparameters that can be adjusted meaningfully, similar to adjusting the focus of a microscope, with different values bringing different aspects of the data into view \citep{diaz2021review}. For the methods based on a $k$NN graph, one such parameter is the number of neighbors ($k$) or its equivalents, such as perplexity in $t$-SNE \citep{skrodzki2023navigating}. Similarly, moving along the attraction-repulsion spectrum in neighbor-embedding methods can be useful for data exploration \citep{damrich2024visualizing}. Beyond conceptual hyperparameters, many methods also have \emph{technical} hyperparameters (for example, those related to optimization). 
    For most use cases, we recommend leaving these at their default values.

    \item \textbf{Investigate unusual shapes or patterns.}
    Odd and unusual shapes in 2D or 3D embeddings can highlight issues with the data or the preprocessing pipeline, calling for additional quality control and filtering steps (for some examples, see \citet{lause2021analytic,gonzalez-marquez2024, nazari2023geometric}). For example, unusually elongated shapes can be the result of erroneously including sample numbers as a spurious feature.

    \item \textbf{Use features and metadata to color the points.}
    The samples in the embedding can be colored by individual features to reveal which of them are influencing which aspects of the embedding. This can help interpret the embedding in terms of the original features. 
    In addition, individual samples often have associated metadata that are not part of the features used for dimensionality reduction (e.g., sample collection site, sample collection date, various measures of sample quality, etc.).  Coloring an embedding by metadata variables can identify problematic batch effects or highlight unexpected findings (for examples of the latter, see \citealt{gonzalez-marquez2024}).

\begin{figure}[t]
    \centering
    \includegraphics[width=\linewidth]{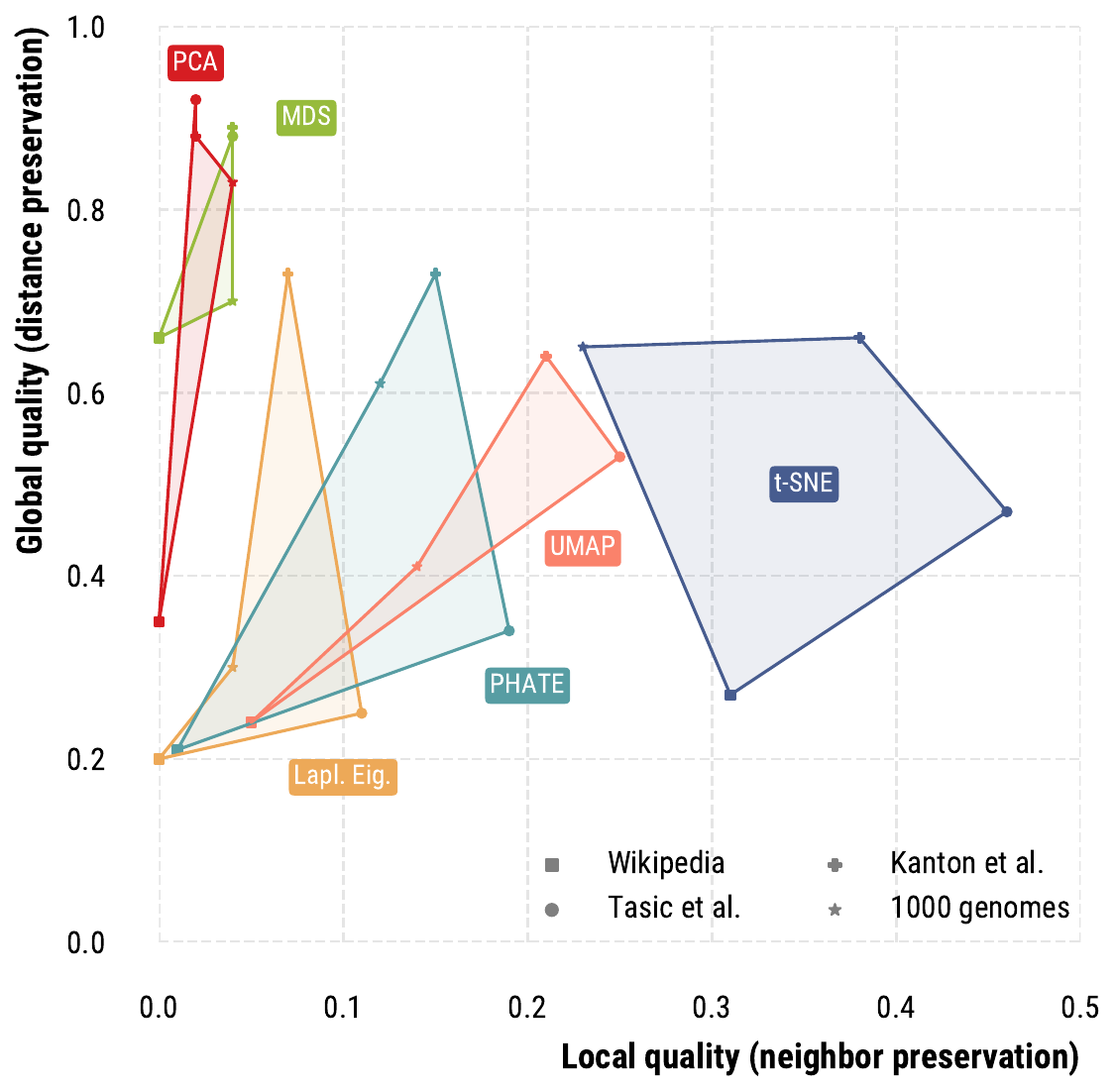}
    \caption{All embeddings shown in Figures~\ref{fig:showcasing_wikipedia}--\ref{fig:showcasing_genomes}, evaluated with two metrics: the fraction of 10 nearest neighbors preserved (local quality) and the correlation between high-dimensional and two-dimensional pairwise distances (global quality).}
    \label{fig:metrics}
\end{figure}

    \item \textbf{Be aware of method limitations.}
    When using an embedding method, it is important to be aware of its limitations and possible distance distortions (see \Cref{subsec:eval} for details). For example, points overlapping in a 2D or 3D PCA projection or Laplacian eigenmaps may be highly distinct and separated along subsequent components (\Cref{fig:showcasing_tasic,fig:showcasing_genomes}), and it is common in data exploration to plot pairwise combinations of subsequent components. Spectral embeddings and MDS can exhibit shapes such as horseshoes that are not scientifically meaningful~\citep{diaconis2008horseshoes,morton2017uncovering}.
    Notably, in neighbor embeddings, between-cluster distances are not indicative of the high-dimensional between-cluster distances.

    \item \textbf{Do not perform downstream analysis on 2D embeddings.}
    2D embeddings are not suited for downstream computational analysis, as they can introduce distortions and artifacts that will be picked up downstream. It is usually more appropriate to perform regression, classification, or clustering on  higher-dimensional data, and only use 2D embeddings for exploration and communication. See \Cref{sec:challenges} about using embeddings of higher dimensionality (e.g., 5D or 10D) for downstream analysis.  

    \item \textbf{Always independently test your hypotheses.}
    As we emphasize throughout the paper, 2D and 3D embeddings are an exploratory tool that can reveal unexpected patterns, suggest hypotheses, and guide researchers toward a confirmatory analysis. However, it is important to further verify such observations.
    A 2D or 3D embedding should not be used to support definitive scientific statements about the data. Rather, it can serve as a starting point for further analysis. 
\end{enumerate}

\paragraph{Visualization}

Here, we note some useful hints and possible pitfalls when preparing figures with 2D embeddings. 

\begin{enumerate}
    \setcounter{enumi}{11}

    \item \textbf{Choose an appropriate color map.}
    Patterns in data can jump out much more clearly given a suitable color contrast, and inadequate coloring can distort the perception of patterns in the data. Embeddings are frequently colored by discrete class labels (\Cref{fig:showcasing_wikipedia,fig:showcasing_tasic,fig:showcasing_kanton,fig:showcasing_genomes}), which can become challenging if there are many classes. Specialized packages exist for generating extended discrete color palettes \citep{mcinnes_glasbey, neuwirth_rcolorbrewer_2022} as well as color-blind-friendly palettes \citep{steenwyk_ggpubfigs_2021, rocchini_under_2024}. 
    Continuous metadata or features should be encoded using continuous color maps (for some examples, see \citet{huisman2017brainscope, diaz2019umap}). 
    It is important to ensure that these color maps are \emph{perceptually uniform}\,---\,i.e., that equal steps in the colormap reflect equal steps in the data \citep{liu_somewhere_2018}.
    When continuous feature values have a clear midpoint (e.g., 0) and the deviations from this midpoint are of interest, diverging color maps should be used.

    \item \textbf{Avoid overplotting.}
    Scatter plots with too many points can become crowded and difficult to parse. Reducing the size and increasing the transparency of points can highlight patterns that are otherwise obscured. 
    When classes have different sizes, smaller groups may get hidden underneath larger ones in the visualization, which can be mitigated by plotting larger groups first. Sometimes, groups in the data can also be hidden in a scatter plot due to an unfortunate row order in the data matrix. Consider randomizing the plotting order of points.
    
    \item \textbf{Remove axis ticks and labels.}
    While for some methods, e.g., MDS or diffusion maps, Euclidean distances in the embedding space are meaningful, for methods that do not preserve distances, such as neighbor embeddings, embedding distances are not interpretable. Tick marks suggest that distances between points or clusters can be measured, so it is better to remove them. 
    Labeling the axes as ``PC1 (13\%)'' and ``PC2 (10\%)'' (showing the PC number and the fraction of explained variance) makes sense for PCA embeddings. However, this is not the case for MDS or neighbor embeddings: These methods are invariant to rotation, such that no separate meaning can be attached to the individual axes. Hence, one may want to drop axis labels such as ``t-SNE 1'' and ``t-SNE 2'' entirely. Square axis frames can also be replaced with circular frames \citep{nonato2018multidimensional}.
    
    \item \textbf{Fix the aspect ratio.}
    Squeezing or stretching the embedding in one direction distorts the distances between points and should be avoided. In most cases, embeddings should be plotted in 1:1 aspect ratio, which is usually not the default plot setting and needs to be manually set when producing a figure (in Python, via \texttt{plt.axis("equal"))}. This is true not only for neighbor embeddings but also for PCA, where each component naturally has different variance (see \Cref{fig:turtles}). 
    An alternative approach often used in PCA and factor analysis (and especially in biplot visualizations, see \citealt{gabriel1971biplot}) is to display standardized components, with each component scaled to have unit variance. 

    \item \textbf{Be intentional and upfront with aesthetic choices.}
    Figure design and aesthetic choices can guide readers toward correct interpretations. For example, if a published embedding should serve as a map, serif fonts or map-like color palettes can invoke viewing habits familiar from geographical maps (with all associated conventions around distortions etc.), rather than those of classical statistical graphics (for an example, see \citealt{manno2021molecular}).
\end{enumerate}

\paragraph{Communication}

Here, we highlight some key considerations for presenting embeddings in scientific papers.

\begin{enumerate}
    \setcounter{enumi}{16}

    \item \textbf{Emphasize method limitations.}
    Be clear in communicating what interpretations and conclusions can be drawn from the embedding. For example, when showing a neighbor-embedding plot with clearly separated clusters, emphasize that between-cluster distances and placement of the clusters with respect to each other may not be meaningful. If an embedding plays a big role in your narrative, report evaluation measures such as the variance explained or the fraction of preserved nearest neighbors (\Cref{subsec:eval}).
    
    \item \textbf{Report details to ensure reproducibility.}
    When reporting your methods, state the versions and parameters of the software required to reproduce your results.
    For example: ``We used the Python implementation of UMAP \texttt{umap-learn} version 0.5.0 with default minimum distance and \texttt{n\_neighbors} set to 50''.
    Making your code available in a public repository and listing the dependencies and their versions following community standards is recommended to facilitate reproducibility.
    
    \item \textbf{Be transparent about data exploration.}
    If subsequent analyses were carried out as a result of embedding-guided data exploration, state this in the Methods section. If a range of methods or parameter choices was used in the exploration process, mention this in the text as well: ``We varied the values of hyperparameter \texttt{X} from 10 to 50, obtaining qualitatively similar results.''

    \item \textbf{Use supplementary embeddings.}
    Different methods or hyperparameters can highlight different aspects of data structure (see \Cref{item:consider-hyperparameters,item:multiple-embeddings} above). Providing additional embeddings as supplementary materials can improve readers' understanding of the data without overburdening the main text. 
\end{enumerate}

\section{Challenges}
\label{sec:challenges}

In this section, we discuss important open questions and challenges for the field of low-dimensional embeddings. We group these challenges into three categories: related to algorithm development, related to human interpretation and interaction, and related to the evaluation of dimensionality-reduction outputs. 

\subsection{Algorithm development}

\paragraph{Scalability}

Many recent algorithmic developments have been motivated by the ever-growing size of available datasets, and we expect this trend to continue in the future. More work is needed on efficient GPU implementations \citep{chan2018t,pezzotti2019gpgpu,nolet2021bringing} and parallelization across multiple GPUs. Another promising direction is compressing the data by reducing numerical precision or converting the data to binarized form \citep{zhang2021faster, bouland2023consequences}. This will likely lead to a trade-off between embedding quality and the runtime, with optimal choices depending on the application.

Another scaling possibility lies in downsampling methods that construct an embedding based on \textit{landmark} points \citep{loukas2019graph, pezzotti2016hierarchical, silva2002global,long2019landmark, moon2019visualizing,shen2022scalability}. Coarse-graining methods that maintain the topological and geometric structure of the data and avoid introducing artifacts~\citep{Huguet23a, loukas2019graph} deserve more research attention. 
Furthermore, additional efforts could be devoted to developing interactive hierarchical embeddings \citep{pezzotti2016hierarchical, marcilio2024humap} that load and/or construct a more detailed embedding on zoom-in. A related approach is to construct an initial embedding using a subset of the data, and then progressively refine the embedding as additional samples are coming in, e.g., in an environment with streaming data \citep{pezzotti2016approximated}. 

\paragraph{Optimization}

While spectral methods rely on eigendecomposition and produce a solution corresponding to a global minimum of their loss function, many popular embedding algorithms (including the neighbor-embedding methods $t$-SNE and UMAP) rely on the iterative optimization of non-convex loss functions (Section~\ref{sec:algorithms}), and hence, can suffer from local minima. Optimization heuristics such as informative initialization \citep{kobak2021initialization} are employed to alleviate this problem. Further research is needed to develop methods that combine convex optimization of spectral techniques with neighbor embeddings.

Multiscale methods that preserve both local and global structures could help here as well, as reproducing global patterns constrains the overall shape of the embedding and leaves less freedom for local structures to move around \citep{lee2024forget}. 
Another promising optimization approach is dimensionality annealing or, alternatively, enabling embedding points to move along temporarily added dimensions \citep{vladymyrov2019no}. 

\paragraph{Non-standard metrics and metric learning}

Most methods reviewed here rely on a particular distance metric, such as Euclidean or cosine. Such metrics may not be optimal for heterogeneous data combining numerical and categorical features, in the presence of missing data \citep{cdb2018drnap,gilbert2018unsupervised,mishne2019co}, or for data with rich internal symmetries, such as text or image datasets. 
Developing embedding methods that can learn adequate distance metrics in such cases, revealing otherwise hidden data structures, remains an important challenge~\citep{lin23b}. Such metric learning can be guided by supervised signals \citep{rhodes2021random, rhodes2023gaining} or by self-supervised learning approaches. Recently, self-supervised learning based on data augmentation has been used to optimize a neural network implementing 2D embedding of image data~\citep{bohm2022unsupervised}.

\paragraph{Optimal dimensionality}

In this review, we focused on low-dimensional embeddings for data exploration, which typically demands very few dimensions. Another goal of low-dimensional embeddings may be downstream analysis, such as clustering, which may be challenging to do in the original high-dimensional space. For example, spectral clustering~\citep{ng2001spectral,shi2000normalized} applies $k$-means clustering to Laplacian eigenmaps. In this case, the appropriate embedding dimensionality is typically larger than two. Some recent work obtained promising clustering results working with neighbor embeddings of dimensionality 5--10 \citep{grootendorst2022bertopic,diaz-papkovich_topological_2023}. More work is needed to benchmark and to study the theoretical properties of such approaches. A related challenge is to identify an optimal embedding dimensionality, perhaps related to the intrinsic dimensionality of the data \citep{levina2004maximum,camastra2016intrinsic}.

\subsection{Interpretability and interactivity}

\paragraph{User studies}

As low-dimensional embeddings are tools for visual data analysis, it is important to understand how people use such embeddings in practice. Nevertheless, user studies have been relatively rare. 
The few existing studies showed, inter alia, 
that the most effective embedding technique depends on the task \citep{xia2021revisiting}, 
that people tend to disregard the methods' mathematical objectives and use them as black boxes \citep{Morariu_TVCG2023}, and that experts and novices evaluate embeddings differently \citep{Lewis_AMCS2012}: novices prefer to see more continuous gradual structures, whereas experts adapt their preference for gradual vs. clustered structures depending on the dataset. Also, people prefer to use embeddings that ``catch the eye'', which, in turn, is influenced by class separability \citep{doh2025understanding}. However, larger and more focused user studies are needed, with participants facing specific data-exploration goals.

\paragraph{Interpretability}

Due to the nonparametric nature of many embedding algorithms, low-dimensional positions may be difficult to interpret and to relate to the original variables \citep{bibal2019measuring}. Questions like ``Which original features cause these two clusters to be separated?'' or ``Why is this outlier so far away?'' are common, and answering them requires follow-up analysis. Explainability tools can make these questions easier to tackle. Even for parametric but highly nonlinear methods, such as autoencoders, interpretation can be challenging. 

Several studies proposed post-hoc explainability approaches, fitting intrinsically interpretable models like linear regressions or decision trees to relate embedding coordinates to the original features in local regions of the embedding space \citep{bibal2020explaining, lambert2022globally,ovcharenko2024feature}.
Nonlinear compositions of functions from a user-defined dictionary were also used for this purpose \citep{koelle2022manifold, koelle2024consistency}. 
Other works enriched the visualizations with additional graphical elements, annotating regions of the embedding space~\citep{Kandogan2012, Pagliosa2016, Tian2021, novak2023framework, policar2024vera}. 
Moreover, several works developed intrinsically interpretable approximations of existing embedding methods \citep{bibal2021biot, couplet2023natively}. Compared to post-hoc explainability techniques, such approaches can enable interpreting the embedding directly, rather than through local approximations. More work is needed in this direction, in particular to integrate interpretability measures into the design of dimensionality-reduction algorithms. 

\paragraph{Post-hoc validation}

Dedicated techniques are needed to enable users to confirm the existence of patterns they see in the low-dimensional embedding in the high-dimensional space. This would allow users to identify real patterns and not over-interpret spurious ones appearing in the visualization due to method limitations.
For instance, computational-topology tools such as persistent homology can validate the presence of loops \citep{edelsbrunner2002topological,damrich2023persistent,kohli2024rats}.
The clusters visible in an embedding should become similarly verifiable, potentially by other means. While modern neighbor embeddings excel at representing clusters in the data, the clustering illusion is a recognized cognitive bias \citep{Gilovich1991}, and sometimes an embedding can be misleading in suggesting the presence of clusters.

\paragraph{Interactive tools}

Data exploration and communication can be enhanced through software that allows dynamic interaction with embeddings \citep{sacha2016visual,Sacha2017}. This includes capabilities such as zooming, displaying metadata on mouse-over events, providing summaries for selected points, dynamic switching between multiple embeddings of the same data or showing paired embeddings side-by-side \citep{huisman2017brainscope}, interactive visualizations of feature gradients \citep{li2023spacewalker}, animations illustrating how embeddings change over time or in response to hyperparameter adjustments~\citep{damrich2024visualizing}, and many more.
Motion can be particularly effective at communicating subtle changes, serving as an important perceptual channel \citep{ware_motion_2006,heer_animated_2007,Munzner2014}, and
interactivity can identify patterns that may be obscured in static images \citep{pezzotti2016approximated}.
Interactive tools have also been used to identify distortions in the embedding~\citep{LespinatsAupetit2011,Heulot2013} as well as to explain structures in the embedding space in relation to the original high-dimensional features~\citep{Stahnke2016,bibal2021ixvc,Eckelt2022}.

There are ongoing efforts to develop hierarchical data representations, where users can select a subset of points with a lasso-like tool and then generate a new embedding using only the selected points to study finer-scale substructure \citep{tino2001principled, pezzotti2016hierarchical, van2017visual, marcilio2024humap, hollt2019focus+}.
This has led to the development of software such as ManiVault \citep{vieth2023manivault}, which allows rapid prototyping of dedicated data viewers, Cytosplore \citep{hodge2019conserved}, a viewer developed for single-cell and spatial data, or DeepScatter \citep{deepscatter}, an in-browser interactive viewer supporting data loading on demand.

\subsection{Evaluation}
\label{subsec:eval}

\paragraph{Quality measures}

Dimensionality-reduction methods produce outputs that can be difficult to assess quantitatively \citep{machado2025necessary}, and multiple complementary measures have been suggested to evaluate the quality of a given embedding (Section~\ref{sec:showcasing},  Figure~\ref{fig:metrics}).

The preservation of pairwise distances can be assessed through stress-based measures \citep{borg1997modern}, Shepard diagrams \citep{leeuw2014shepard}, and normalized measures such as $\sigma$-distortion  \citep{chennuru2018measures}. 
Local distance preservation can also form the basis of global distortion measures \citep{jang2021riemannian}. 
To measure the preservation of neighborhoods across various scales, from local to global, ranked distances are often used \citep{france2007development, lee2009quality, venna2010information, cdb2022fastms, griparis2016dimensionality, novak2023framework}. If ground-truth class labels are available, then additional measures can be based on classification accuracy, on class separation (e.g., silhouette score, \citet{rousseeuw1987silhouettes}), or on the extent that clustering in the embedding reflects the original classes  \citep{mokbel2010effect, mokbel2011quality,espadoto2019toward,huang2022towards,wang2023comparative,lause2024art}. If the high-dimensional data contains non-trivial topological structures, persistent homology can be used to evaluate their preservation in the embedding \citep{rieck2015persistent, paul2017study}.

These criteria are useful to compare the performance of several dimensionality-reduction methods, but their absolute values can be difficult to interpret directly. As the theoretically best performance achievable on a given dataset is typically unknown, the question of whether an embedding is \textit{good enough} often remains unanswered. 
Therefore, developing model-independent and robust measures that produce interpretable absolute values remains an important research direction.

\paragraph{Fine-grained visualization of embedding quality}

Embedding quality can vary across the embedding space, and local quality measures can provide additional information compared to a single global measure computed over the entire embedding. It is important to develop software that enables fine-grained evaluations and integrates them into visualizations, e.g., by coloring points according to the local embedding quality \citep{pezzotti2016approximated, thijssen2024interactive, tian2023measuring, benato2023measuring}, or by overlaying the embedding with additional graphical elements~\citep{Aupetit2007,Seifert2010,Schreck2010,Heulot2013,Martins2014}.

\paragraph{Theoretical guarantees}

For many embedding methods, it is difficult to establish rigorous theoretical guarantees, and despite some progress in this direction \citep{tenenbaum2000global, zhang2004, linderman2022dimensionality, cai2022theoretical, arora2018analysis}, further work is needed. Such guarantees can, for example, include bounds on the distortion incurred in the low-dimensional embeddings, or thresholds on the noise levels and inter-cluster linkage probabilities that allow recovery of the true clusters from their noisy observations. Even simple methods with a long history, such as classical MDS, can have surprising mathematical properties \citep{lim2024classical, kroshnin2022infinite}. Closer theoretical connections between different visualization methods~\citep{noack2009modularity, damrich2023from, huguet2024heat} will clarify the relative strengths of each method. Further advancements in theoretical understanding could also provide a more rigorous framework for evaluating the performance of dimensionality-reduction methods.

\section{Conclusion and vision}
\label{sec:vision}

Today, researchers live in a world where their capacity to access or produce data often exceeds their ability to understand it.
Since much of this data is high-dimensional, machine-learning methods that produce low-dimensional embeddings have gained importance in exploratory data analysis. 
In this paper, we argued that low-dimensional embeddings and visualizations can guide analyses and discussions, highlight interesting patterns, and yield new hypotheses, investigations, or questions. By creating compelling figures and adding transparency to the scientific process, embeddings also play an important role in research communication.  
Paraphrasing George Box \citep{box1979robustness}: \emph{All embeddings are wrong, but some are useful.}

As data availability continues to improve, we anticipate an increased interest in low-dimensional embedding methods across various data-intensive disciplines. This includes not only machine learning and biotechnology, but also interdisciplinary fields like digital humanities and computational social science. The continued cross-pollination between methods and applications across such widely different domains promises exciting methodological developments for the field. Research on low-dimensional embeddings has already made tremendous progress (compare Figure~\ref{fig:turtles} with Figure~\ref{fig:showcasing_wikipedia}), and we expect further advances in the oncoming years.

However, as we emphasized throughout the paper, embedding methods often involve technical complexities that may not be immediately apparent to their users. Making these methods accessible to interdisciplinary researchers requires promoting awareness of best practices and methodological limitations. It also requires addressing the many remaining algorithmic and interpretational challenges identified above. Overall, embedding methods hold strong potential to shape the future of data-driven research.

\clearpage

\section*{Acknowledgments}
This work was conceived at the Dagstuhl seminar 24122 supported by the Leibniz Center for Informatics. 
CdB conducted part of this work while being a beneficiary of an FSR Incoming Post-doctoral Fellowship from UCLouvain. 
ADP is supported by National Institutes of Health (NIH) Grant R35 GM139628. 
MB is funded by the Deutsche Forschungsgemeinschaft (DFG, German Research Foundation) under Germany's Excellence Strategy EXC 2181/1 - 390900948 (the Heidelberg STRUCTURES Excellence Cluster). 
KB is supported by the Netherlands Organisation for Scientific Research (NWO) under Vidi grant number VI.Vidi.193.098. 
CC conducted part of this work while supported by Digital Futures at KTH Royal Institute of Technology. 
SD is supported by the German Ministry of Science and Education (BMBF) via  the T\"{u}bingen AI Center (01IS18039) and by the National Institutes of Health (UM1MH130981). 
SD and DK are supported by the Gemeinn\"{u}tzige Hertie-Stiftung. 
E\'AH is supported by the National Science Foundation (NSF CAREER Grant IIS-1943506). 
JAL is a research director with the Belgian F.R.S.-FNRS (Fonds National de la Recherche Scientifique). 
BR acknowledges that this work has received funding from the Swiss State Secretariat for Education, Research and Innovation (SERI). 
GW is supported by a Humboldt Research Fellowship, CIFAR AI Chair, and NSERC Discovery grant 03267. 
GM is partially supported by the National Science Foundation grants CCF-2217058 and EFRI BRAID 2223822. 
DK is a member of the Germany’s Excellence cluster 2064 ``Machine Learning --- New Perspectives for Science'' (EXC 390727645). 
The content provided here is solely the responsibility of the authors and does not necessarily represent the official views of the funding agencies. 

\bibliographystyle{plainnat}
\bibliography{main.bib}

\clearpage

\section{Methods}
\label{sec:methods}

\paragraph{Data and code availability}

All analysis code is openly available at \url{https://github.com/dkobak/low-dim-embeddings-review/}. All datasets are openly available and, for convenience and reproducibility, the preprocessed versions are shared in our repository.

\paragraph{Data sources and preprocessing} 

For all datasets, we aimed to use standard preprocessing steps, usually taken from prior work.

\begin{itemize}

\item
The Simple English Wikipedia data were downloaded from \url{https://huggingface.co/datasets/Cohere/wikipedia-22-12-simple-embeddings}. This dataset contains 485\,859 text paragraphs from the Simple English Wikipedia, represented as 768-dimensional vectors using a language model (\texttt{multilingual-22-12} from Cohere.AI; the model generated embeddings of text paragraphs with the prepended Wikipedia article title). We scaled each of these vectors to have a unit $l_2$-norm, ensuring that Euclidean and cosine distances yield identical $k$NN sets.

To annotate the UMAP embedding, we used the Toponymy library (\url{https://github.com/TutteInstitute/toponymy}). Toponymy performed HDBSCAN clustering of the 2D UMAP embedding (with \texttt{cluster\_selection\_method="leaf"}). The cluster labels were generated by Toponymy through extracting various types of information from each cluster (including representative exemplars, distinguishing keyphrases, and relevant sub-clusters) and passing this to a generative LLM (\texttt{c4ai-command-r-08-2024} from Cohere.AI) to generate a concise human-readable name.

\item
Tasic et al. data \citep{tasicSharedDistinctTranscriptomic2018} contain 23\,822 cells from mouse cortex, with 45\,768 genes as features. This is an integer count matrix showing how many RNA molecules of each gene were detected in each cell. We followed preprocessing steps used in \citet{kobak2019art}. This includes selecting 3\,000 highly variable genes, normalizing by the total gene count in each cell, $\log(1+x)$ transformation, and finally projecting to 50 components via PCA. The remaining 50 components preserve 62.6\% of the total 3\,000-dimensional variance. Note that the fraction of variance explained by the first two PCs reported in the main text is with respect to the 50-dimensional data, and all quality measures for all methods were computed based on the 50-dimensional data as well. 

\item
Kanton et al. data \citep{kantonOrganoidSinglecellGenomic2019} contain 20\,272 cells from human brain organoids (cell line 409b2), with 32\,856 genes as features. We followed preprocessing steps used in \citet{bohm2022attraction} and \citet{damrich2024visualizing}. As above, these include selecting 1\,000 highly variable genes, normalization, log-transformation, and PCA projection to 50 components (preserving 60.2\% of the total 1\,000-dimensional variance). As above, the fraction of variance explained by the first two PCs reported in the main text is with respect to the 50-dimensional data, and all quality measures were computed based on the 50-dimensional data as well.

\item
The raw 1000 Genomes Project data \citep{global_2015} are available at \url{https://ftp.1000genomes.ebi.ac.uk}.
This dataset contains 3\,450 human genotypes.
We used PLINK \citep{purcell_plink_2007} to filter for
linkage disequilibrium and high variability regions as described in \citep{diaz2019umap} and \citep{diaz-papkovich_topological_2023}. The specific PLINK parameters are given in the shell scripts available in our GitHub repository. This resulted in an integer-valued data matrix with 53\,999 features, containing values 0, 1, and 2, representing the number of alleles differing from a reference genome. We replaced missing values, coded as $-1$, by 0. Preprocessing did not involve PCA projection.
\end{itemize}

\paragraph{Algorithm implementations} 

For all methods, we aimed at using the most standard Python implementation with default parameters.

\begin{itemize}
    \item For PCA, we used \texttt{sklearn.decomposition.PCA} from \texttt{scikit-learn} version 1.5.1 \citep{pedregosa2011scikit}, with \texttt{svd\_solver=`arpack'}.
    
    \item For MDS, we used the stochastic SQuadMDS algorithm \citep{lambert2022squadmds} with the default parameters (including PCA initialization) using the code provided at \url{https://github.com/PierreLambert3/SQuaD-MDS-and-FItSNE-hybrid} (commit \texttt{4cfa8b4}). The 1000 Genomes dataset was small enough to use a non-stochastic MDS implementation, so for that dataset, we used \texttt{sklearn.manifold.MDS} with the default parameters and manually provided PCA initialization to the \texttt{fit()} function.

    \item For Laplacian Eigenmaps, we used \texttt{sklearn.manifold.SpectralEmbedding}  with \texttt{n\_neighbors=100}. We tried various numbers of nearest neighbors and obtained similar embeddings. The default value in this implementation is 10\% of the sample size, which is computationally unfeasible for large datasets. For the Simple English Wikipedia dataset, we used \texttt{solver=`amg'} as the default solver only works with dense matrices and raised memory errors. 
    
    \item For PHATE, we used the \texttt{phate} library \citep{moon2019visualizing}, version 1.0.11, with all default parameters (in particular, the number of nearest neighbors was set to~5).
    
    \item For $t$-SNE, we used the \texttt{opentsne} library \citep{polivcar2019opentsne}, version 1.0.2, with all default parameters (in particular, perplexity~30 and PCA initialization). 
    
    \item For UMAP, we used the \texttt{umap-learn} library \citep{mcinnes2018umap}, version 0.5.7, with all default parameters (in particular, 15~nearest neighbors and spectral initialization). For the Simple English Wikipedia dataset, we increased the number of training epochs to 500 to ensure better convergence. 
\end{itemize}

Some embeddings were flipped horizontally and/or vertically to ease visual comparison between methods.

\paragraph{Quality assessment} 

We employed four quality measures to evaluate the 24 embeddings (four datasets and six methods). 
These quality measures are described below, with $n$ denoting the sample size in a given dataset, $\delta_{ij}$ denoting the high-dimensional Euclidean pairwise distances, and $d_{ij}$ denoting the low-dimensional Euclidean pairwise distances. 

\begin{itemize}
    \item The Pearson correlation between $\delta_{ij}$ and $d_{ij}$ \citep{becht2019dimensionality,kobak2019art}. The correlation was computed across the set of all $n(n-1)/2$ pairwise distances. For the Wikipedia dataset, we estimated the correlation using a random subset of 1\,000 points, computing the correlation over all $1\,000\cdot(n-1)$ pairwise distances between these points and all points in the dataset. 
    This measure serves as the $y$-axis in Figure~\ref{fig:metrics}. 
    
    \item The $\sigma$-distortion \citep{chennuru2018measures} between $\delta_{ij}$ and $d_{ij}$. This measure is defined via the ratios $\rho_{ij} =d_{ij} / \delta_{ij}$. If we denote the average ratio as $\bar \rho$, then the $\sigma$-distortion is defined as the variance of the normalized ratios $\rho_{ij} / \bar\rho$, i.e., the average value of $(\rho_{ij} / \bar\rho - 1)^2$. The variance is computed over all $n(n-1)/2$ normalized ratios.    
    Note that this measure is not affected by the scale of the embedding distances $d_{ij}$.
    For the Wikipedia dataset, we estimated the $\sigma$-distortion using a random subset of 1\,000 points and computing the variance over $1\,000\cdot(n-1)$ normalized ratio values.
    
    \item The average overlap between the 10 nearest neighbors in the high-dimensional space and the 10 nearest neighbors in the embedding, averaged across all data points \citep{lee2009quality,kobak2019art}. This measure was computed using exact nearest neighbors, averaging across all $n$ points in the dataset. This measure serves as the $x$-axis in Figure~\ref{fig:metrics}.
    
    \item The area under the rescaled neighbor-preservation curve \citep{lee2015multi}. This measure quantifies the reproduction of neighborhoods across neighborhoods of varying size. The computation is based on the average overlap between high-dimensional and low-dimensional neighborhoods of size $k\in[1, n-2]$, 
    \begin{equation}
        Q(k) = \frac{1}{n} \sum_{i} \frac{\left|\nu_i^k \cap n_i^k\right|}{k},
        \label{eq:qnx}
    \end{equation}
    where $\nu_i^k$ and $n_i^k$ are sets of $k$ nearest neighbors of data point $i$ in the original space and in the embedding, respectively. Note that for $k=10$, this gives the same measure as described above and used in  Figure~\ref{fig:metrics}. For the Wikipedia dataset, we estimated $Q(k)$ as an average over a random subset of 1\,000 points, based on determining, for each point in the subset, the exact high-dimensional and low-dimensional neighborhoods among all points in the dataset, for all values of ${k\in[1, n-2]}$ \citep{novak2023framework}. 
    
    For a random embedding, the expected value of $Q(k)$ is $k/(n-1)$. Following \citet{lee2015multi}, we rescale $Q(k)$ as 
    \begin{equation}
        R(k) = \frac{(n-1)Q(k)-k}{n-1-k}\;,
        \label{eq:rnx}
    \end{equation}
    such that for any value of $k$, $R=0$ corresponds to the chance level and $R=1$ corresponds to a perfect neighborhood overlap. 
    
    Finally, we compute the area under the $R(k)$ curve from $k=1$ to $k=n-2$, with logarithmic scaling of the $k$-axis to give more weight to smaller neighborhoods (and with a scaling factor to make AUC lie between 0 and 1), as 
    \begin{equation}
        \text{AUC}=\Big(\sum_{k=1}^{n-2}k^{-1}\Big)^{-1} \sum_{k=1}^{n-2}\frac{R(k)}{k}\;.
        \label{eq:auc}
    \end{equation}
\end{itemize}


\begin{table*}[t]
\caption{Values of the quality measures for all embeddings. The best value in each row is highlighted in bold. As indicated by the arrow annotations to the quality measures ($\uparrow|\downarrow$), for $\sigma$-distortion, lower scores are better, whereas for all other measures, higher scores are better.}
\label{table:qa}
\centering
\begin{tabular}{ccrrrrrr}
\toprule
Dataset & Quality measure & PCA & MDS & LE & PHATE & $t$-SNE & UMAP \\
\midrule
\multirow{4}{*}{Tasic et al.}& Distance correlation $\uparrow$& \textbf{.92} & .88 & .25 & .34 & .45 & .53 \\
&$\sigma$-distortion $\downarrow$& .34 & \textbf{.13} & 1.03 & .23 & .19 & .20 \\
&$k$NN preservation $\uparrow$& .02 & .04 & .11 & .19 & \textbf{.46} & .25 \\
&AUC $\uparrow$& .21 & .29 & .32 & .34 & \textbf{.50} & .41 \\
\midrule
\multirow{4}{*}{Kanton et al.}& Distance correlation $\uparrow$& .88 & \textbf{.89} & .73 & .72 & .67 & .64 \\
&$\sigma$-distortion $\downarrow$& .21 & \textbf{.11} & .40 & .35 & .15 & .20 \\
&$k$NN preservation $\uparrow$& .02 & .04 & .07 & .14 & \textbf{.38} & .21 \\
&AUC $\uparrow$& .24 & .29 & .26 & .31 & \textbf{.46} & .37 \\
\midrule
& Distance correlation $\uparrow$& \textbf{.83} & .70 & .30 & .65 & .68 & .46 \\
1000 Genomes& $\sigma$-distortion $\downarrow$& .39 & \textbf{.21} & .91 & .28 & .23 & .36 \\
Project&$k$NN preservation $\uparrow$& .04 & .04 & .04 & .12 & \textbf{.20} & .14 \\
&AUC $\uparrow$& .24 & .20 & .20 & .29 & \textbf{.41} & .27 \\
\midrule
& Distance correlation $\uparrow$& .32 & \textbf{.63} & .18 & .14 & .27 & .23 \\
Simple English& $\sigma$-distortion $\downarrow$& .28 & \textbf{.19} & 1.88 & .80 & .22 & .27 \\
Wikipedia& $k$NN preservation $\uparrow$& .00 & .00 & .01 & .02 & \textbf{.26} & .08 \\
& AUC $\uparrow$& .06 & .07 & .08 & .07 & \textbf{.21} & .13 \\
\bottomrule
\end{tabular}
\end{table*}

\end{document}